\documentclass{article}
\usepackage[preprint]{neurips_2026}
\usepackage[utf8]{inputenc}
\usepackage[T1]{fontenc}
\usepackage{hyperref}
\usepackage{url}
\usepackage{booktabs}
\usepackage{amsfonts}
\usepackage{amssymb}
\usepackage{amsmath}
\usepackage{nicefrac}
\usepackage{microtype}
\usepackage{xcolor}
\usepackage{graphicx}
\usepackage{multirow}
\usepackage{array}
\usepackage{tabularx}
\usepackage{subcaption}
\usepackage{float}
\usepackage{placeins}
\usepackage{tikz}
\usetikzlibrary{arrows.meta,positioning,shapes,fit,backgrounds}

\newcommand{\dd}[1]{\texttt{#1}}

\definecolor{darkgreen}{rgb}{0.0,0.45,0.0}
\definecolor{darkred}{rgb}{0.7,0.0,0.0}
\definecolor{midblue}{rgb}{0.1,0.2,0.7}

\title{Cross-Architecture Steering Transfer in Language Models: A Systematic Empirical Study}

\author{%
  Ayushi Agarwal\\
  \small Independent Researcher
}

\begin{document}
\maketitle

\begin{abstract}
Independently trained large language models may develop shared internal
representations of semantic concepts despite architectural differences---but
whether this geometric similarity has functional consequences for cross-model
behavioural control has not been systematically tested.
In this work, we present the first systematic evaluation of whether the shared
geometry of independently trained LLMs is \emph{functionally exploitable} for
cross-model steering transfer, and show that it is, conditionally: concept
directions extracted from one model can steer a different, independently trained
model when sufficient representational capacity has been reached.
We study five open-weight models spanning three parameter scales (0.8B--8B) and
two architectural lineages, training one Sparse Autoencoder per model across 15
semantic domains and testing alignment across all 20 directed model pairs.
Specifically, we observe a suggestive discontinuity near 1.7B parameters:
at $\geq$1.7B scale, 47--49\% of cross-model feature pairs validate
(Pearson $r \geq 0.60$, Procrustes cosines 0.895--0.956), while alignment
degrades sharply below 0.8B despite raw activation geometry being preserved
(with $n{=}1$ model per tier below 7B this cannot yet be confirmed as a
systematic scaling law).
Leveraging this shared structure, cross-model steering vectors (B3-TI) achieve
a 71.0\% win rate across 15 supervised concepts and five models versus 68.0\%
for same-model native vectors; a single universal vector from an unsupervised
shared concept space achieves 67.3\% in 4 of 5 models without any per-model
supervision.
Transfer degrades for models below 1.7B scale and for one model with generation
instability, confirming that functional exploitability of the shared geometry
requires sufficient representational capacity.
Our findings underscore the importance of scale thresholds in mechanistic
interpretability: tools validated at 7B scale may not transfer to smaller models
or frontier architectures without revalidation.
More broadly, we provide the first functional complement to the Platonic
Representation Hypothesis---showing that geometric convergence across
independently trained LLMs is strong enough to support cross-model behavioural
control without fine-tuning, under the identified scale conditions.
\end{abstract}

\section{Introduction}
\label{sec:intro}

A central question in mechanistic interpretability is whether the internal
representations of independently trained language models are \emph{universal}:
not merely similar in the features they respond to, but geometrically aligned in
a way that supports functional operations such as steering, suppression, and
cross-model transfer.
If such shared geometry is not just a structural curiosity but is actually
\emph{exploitable}---if a concept direction from one model can causally control
a different model it has never been trained on---that would mark a fundamental
shift in how we think about cross-architecture interpretability tools.

The Platonic Representation Hypothesis~\citep{huh2024platonic} provides
geometric evidence that representations across modalities and architectures
converge toward a shared statistical model of reality.
Whether this shared geometry has \emph{functional} consequences---specifically,
whether a concept direction extracted from one model can causally steer a
different, independently trained model---remains an open question.
Prior steering-vector work~\citep{turner2023activation,zou2023representation}
demonstrates functional control \emph{within} a single architecture but does not
test cross-architecture transfer.

This work addresses that question in three stages:
\textbf{(1) we establish that cross-architecture feature-level universality
exists} for models at $\geq$1.7B parameters, using SAE-extracted features aligned
across all 20 directed pairs of five independently trained models;
\textbf{(2) we explore whether the shared structure is functionally usable for
steering}, measuring whether concept directions injected at inference time
causally shift output distributions in the expected direction;
\textbf{(3) we demonstrate that cross-model steering transfer works in practice},
with cross-model vectors outperforming naive baselines and a single unsupervised
universal vector generalising to 4 of 5 models without per-model supervision.

Our study makes the following contributions:

\begin{enumerate}
  \item \textbf{Evidence for cross-architecture feature-level universality.}
    Among four models at $\geq$1.7B scale, 47--49\% of MNN-extracted feature
    pairs validate cross-model (Pearson $r \geq 0.60$, Procrustes cosines
    0.895--0.956, co-activation enrichment $28.6\times$ above permutation null),
    with a suggestive discontinuity near 1.7B below which alignment degrades
    substantially ($n{=}1$ model per tier below 7B; systematic scaling requires
    a broader model sweep).

  \item \textbf{Exploration of functional steering transferability.}
    We evaluate whether the shared geometry supports causal concept control
    across architectures, using a bidirectionality criterion
    ($r_{\text{bidir}} = |\Delta^-|/|\Delta^+|$) as evidence that injected
    directions genuinely encode the target concept rather than one-directional
    saturating triggers. Median $r_{\text{bidir}} = 1.71$ for B3-TI vectors
    (82\% of model--concept pairs with $r_{\text{bidir}}\geq 0.5$) confirms T2
    causal evidence.

  \item \textbf{Demonstration that cross-model transfer works.}
    B3-TI cross-model vectors achieve a 71.0\% win rate across 15 supervised
    concepts and five models, compared to 68.0\% for same-model native vectors
    and 65.7\% for a naive zero-projection baseline (the 3.0pp gap is
    directional; paired $t(4){=}1.85$, $p{=}0.14$, underpowered at $n{=}5$ models).
    A single universal vector from an unsupervised shared concept space achieves
    67.3\% across 11 unsupervised concepts in 4 of 5 models without per-model
    supervision.

  \item \textbf{Reproducible pipeline and fully released artefacts.}
    All intermediate and final outputs are released on Hugging Face:
    five per-model residual-stream activation datasets (A2),
    five trained TopK SAEs (A3),
    4,100+ labelled feature files (A4),
    native SAE-decoder and CAA steering vectors per concept (A5),
    20 trained cross-model MLP bridges (B1),
    full pairwise alignment results and statistics (B2),
    B3-TI cross-model vectors for all 20 directed pairs (B3),
    the trained universal Global MLP (C1),
    universal concept cluster mappings and results (C2),
    and 55 C3-Dec / C3-EncDec universal steering vectors (C3),
    together with all evaluation results (D1) and the complete
    steering evaluation dataset.
\end{enumerate}

The paper is organised as follows.
Section~\ref{sec:related} reviews related work.
Section~\ref{sec:method} describes the end-to-end pipeline.
Section~\ref{sec:results} presents results: geometric universality (T1, §\ref{sec:t1}),
domain structure (§\ref{sec:domains}), and functional transfer (T2, §\ref{sec:t2}).
Section~\ref{sec:discussion} discusses scope, failure modes, and future directions.

\section{Related Work}
\label{sec:related}

\citet{bricken2023monosemanticity} showed SAEs decompose residual-stream activations into near-monosemantic features; \citet{gao2024scaling} established TopK SAEs as the scaling-efficient variant; \citet{cunningham2023sparse} evaluated interpretability across GPT-2 scales---we build on this by training TopK SAEs on five architectures and using their features as the cross-model alignment substrate. \citet{turner2023activation} introduced Contrastive Activation Addition (CAA) and \citet{zou2023representation} demonstrated linear representation probes for safety concepts, both operating \emph{within} a single model; we evaluate whether such directions transfer \emph{across} architectures. \citet{kornblith2019similarity} and \citet{nguyen2021wide} established CKA and Procrustes alignment as representational similarity tools; \citet{raghu2017svcca} contributed SVCCA; \citet{huh2024platonic} synthesised these into the Platonic Representation Hypothesis (T1 convergence)---we provide the first T2 functional complement showing T1 alignment is strong enough for zero-shot steering transfer. \citet{haghverdi2018batch} introduced MNN for cross-dataset cell matching in genomics, adapted here for cross-architecture SAE feature alignment; \citet{conneau2018word} applied MNN to cross-lingual embeddings. \citet{wang2024towards} and \citet{lan2024universality} find high structural SAE-feature similarity across LLM pairs under rotation-invariant measures, establishing geometric universality but not testing causal steering transfer; \citet{gurnee2024universal} find 1--5\% of neurons are universal across GPT-2 seeds at neuron granularity, whereas we operate at SAE feature granularity across architecturally distinct models. \citet{lindsey2024crosscoders} introduce sparse \emph{crosscoders} that jointly encode activations from multiple models to produce a shared feature set for model diffing---our Global MLP (C1) is conceptually related but optimises explicitly for cross-model steering transfer via contrastive alignment rather than feature decomposition.

\section{Method}
\label{sec:method}

The pipeline spans eleven steps across three tracks (see Figure~\ref{fig:pipeline}
in Appendix~\ref{app:pipeline} for a visual overview):

\noindent\textbf{Track A\,---\,Single-model} \emph{(run per architecture):}\\
\textbf{A1}~Corpus: 394,508-passage corpus across 15 semantic domains.\\
\textbf{A2}~Activations: residual-stream mean-pool at ${\approx}50\%$ depth (released).\\
\textbf{A3}~SAE training: one TopK SAE per model, 16,384 latents (released).\\
\textbf{A4}~Feature labelling: delta-selection + LLM-judge scoring (4,100+ files released).\\
\textbf{A5}~Native vectors: SAE-decoder and CAA vectors per concept (released).

\noindent\textbf{Track B\,---\,Pairwise cross-model} \emph{(20 directed pairs):}\\
\textbf{B1}~Feature alignment + MLP bridge training via MNN matching (released).\\
\textbf{B2}~Pair validation: Pearson $r\!\geq\!0.60$, Procrustes cosine, enrichment (released).\\
\textbf{B3}~B3-TI cross-model steering vectors via B1 bridge (released).

\noindent\textbf{Track C\,---\,All-model universal} \emph{(all 5 models jointly):}\\
\textbf{C1}~Global MLP: joint encoder/decoder over all five activation spaces (released).\\
\textbf{C2}~Universal concept discovery: HDBSCAN yields 11 canonical concepts (released).\\
\textbf{C3}~Universal vectors: C3-Dec and C3-EncDec, no per-concept supervision (released).

\noindent\textbf{Track D\,---\,Functional evaluation:}\\
\textbf{D1}~Win rate, bidirectionality, and concept-score $\Delta$ across
30 prompts $\times$ 15 concepts $\times$ 9 strengths (all results released).

Full derivations, hyperparameters, and failure analyses are in the appendix;
each paragraph below points to the relevant section.

\paragraph{Models (Table~\ref{tab:models}).}
We study five open-weight decoder-only models spanning three parameter scales
(GPT-2-large 0.8B, Gemma-2-2B 1.7B, three 7B models: LLaMA-3.1-8B, Mistral-7B,
DeepSeek-7B) and two architectural lineages (GPT-2 absolute-positional MHA;
LLaMA-family RoPE variants including standard MHA, GQA, and sliding-window GQA).
LLaMA (Hermes-3, SFT only) is the sole instruction-tuned model; the remaining
four are unmodified base checkpoints, included to test whether fine-tuning
disrupts representational universality (Section~\ref{sec:t1}); it does not.

\begin{table}[h]
\centering
\caption{Models used in this study.}
\label{tab:models}
\small
\begin{tabular}{llccc}
\toprule
Key & HuggingFace ID & Non-emb params & Hidden dim & Target layer \\
\midrule
\dd{gpt2-large}    & \texttt{gpt2-large}                          & 0.71B & 1\,280 & 19/36 (53\%) \\
\dd{gemma}         & \texttt{google/gemma-2-2b}                   & 1.70B & 2\,304 & 13/26 (50\%) \\
\dd{llama}         & \texttt{NousResearch/Hermes-3-Llama-3.1-8B}  & 6.4B  & 4\,096 & 16/32 (50\%) \\
\dd{mistral}       & \texttt{mistralai/Mistral-7B-v0.3}           & 6.4B  & 4\,096 & 16/32 (50\%) \\
\dd{deepseek}      & \texttt{deepseek-ai/deepseek-llm-7b-base}    & 6.0B  & 4\,096 & 15/30 (50\%) \\
\bottomrule
\end{tabular}
\end{table}

\paragraph{A1--A2: Corpus and activations.}
We assemble 394,508 passages from 17 HuggingFace datasets covering 15 semantic
domains: Python code (3 granularity levels), SQL, four mathematics subsets
(GSM8K, MetaMath, MATH-plus, NuminaMath-CoT), creative writing, academic
abstracts, biomedical, legal, news, open-domain QA, and two general prose
sources. For each passage we record the mean-pooled residual-stream hidden state
at the target layer ($\approx$50\% depth), then z-score normalise per dimension
to remove scale disparities of up to $12\times$ across architectures.
Full specifications are in Appendix~\ref{app:corpus}.

\paragraph{A3: Sparse Autoencoder training.}
One TopK SAE~\citep{gao2024scaling} is trained per model: $64\times$ expansion
for GPT-2-large and Gemma (81,920 and 147,456 features respectively), $128\times$
for the three 7B models (524,288 features each). Ghost gradients are used for 7B
models to prevent dead-neuron accumulation. All five SAEs converge to 0\% dead
features; reconstruction losses range from 0.065 (GPT-2-large) to 0.250 (LLaMA).
Per-model hyperparameters and losses are in Appendix~\ref{app:sae}.

\paragraph{A4: Feature selection and domain assignment.}
For each of the 15 supervised domains, the top-150 SAE features are selected
by differential delta $\delta_c[f] = \bar{F}_{\text{pos},c}[f] - \bar{F}_{\text{neg},c}[f]$;
domain assignment is $\arg\max_c |\delta_c[f]|$. This yields 666--883 labelled
features per model (4,100 total); full configuration in Appendix~\ref{app:a4}.

\paragraph{A5: Native steering vectors.}
Two vector types are computed per (model, concept).
\emph{SAE-decoder vectors}: a confidence-weighted sum of the top-3 SAE decoder
columns for the concept's highest-$\delta_c$ features, $\ell_2$-normalised.
\emph{Contrastive Activation Addition (CAA) vectors}~\citep{turner2023activation}:
mean residual-stream difference between 50 positive and 50 negative passages
for the concept, $\ell_2$-normalised.
The cosine similarity between the two types serves as a within-model quality
check before cross-model transfer is attempted.

\paragraph{B1: Cross-model feature alignment.}
For each of the 20 directed model pairs, MNN-selected feature pairs with
composite alignment score $\geq 0.70$ are retained (3,308 validated pairs
total); a 67M-parameter MLP bridge is trained per directed pair for B3-TI.
Per-pair statistics are in Appendix~\ref{app:b1_pairs}.

\paragraph{B2: Alignment validation.}
Each B1 pair is validated using Pearson $r \geq 0.60$, permutation enrichment
(BH-FDR), Cohen's $d$, Spearman $\rho$, CCC, and RSA\@. SAE-free Procrustes
cosines on raw activations provide architecture-independent upper bounds.
Full statistics are in Appendix~\ref{app:b2_full}.

\paragraph{B3: Cross-model steering vectors.}
\emph{B3-TI (Translation Injection)}: the guide's A5 SAE-decoder vector is
encoded to sparse features, projected through the B1 MLP bridge, and decoded
through the target SAE decoder ($\ell_2$-normalised). A \emph{naive baseline}
(dimension-matched, $\ell_2$-normalised) provides a zero-learning-cost
reference. Two discarded B3 variants are described in Appendix~\ref{app:discarded}.

\paragraph{C1: Global MLP.}
Five per-model encoders ($d_{\text{model}} \to 2048 \to 512$-d shared space)
with symmetric decoders are trained jointly using MSE reconstruction loss plus
NT-Xent contrastive loss~\citep{chen2020simclr} ($\tau{=}0.1$), aligning
same-passage representations from different models while preserving per-model
structure. Training runs for 200 epochs on 8$\times$A100-80GB; final
reconstruction loss is 0.013 with 0\% dead concept-space neurons.
Architecture and training details are in Appendix~\ref{app:c1}.

\paragraph{C2: Universal concept discovery.}
The 4,100 A4-labelled SAE feature vectors are projected into C1 concept space,
dimensionality-reduced with UMAP ($512 \to 30$-d, reducing noise from 58\% to
15.8\%), and clustered with HDBSCAN, yielding 113 clusters spanning $\geq$2
model architectures. Each cluster is LLM-labelled (Claude, temperature~0) and
human-reviewed; after deduplication and noise removal, \textbf{11 canonical
universal concepts} with 5/5 model coverage emerge.
Full cluster tables and canonical mapping are in Appendix~\ref{app:c2}.

\paragraph{C3: Universal steering vectors.}
\emph{C3-Dec}: the C2 cluster centroid is decoded through C1's per-model
decoder and SAE decoder to native hidden-space; polarity resolved by
sign-agreement with A5 CAA vectors.
\emph{C3-EncDec}: the guide's A5 vector is encoded through C1's guide encoder,
decoded through the target's C1 decoder and SAE, and averaged over all guide
models (polarity correct by construction).
Both require no supervision, yielding $11{\times}5 = 55$ vectors per type at
$O(n)$ extension cost versus $O(n^2)$ for B3-TI.

\paragraph{D1: Functional evaluation.}
All six vector types are evaluated under a shared protocol: 30 prompts,
nine-point strength sweep ($s \in \{-5,-3,-2,-1,0,1,2,3,5\}$), additive
residual-stream injection, and DeBERTa-v3-large NLI concept scoring~\citep{laurer2023deberta}.
Primary metrics: win rate ($\Delta > 0$ fraction) and bidirectionality
$r_{\text{bidir}} = |\Delta^-|/|\Delta^+|$; outputs with repetition $>$0.40 excluded.
\dd{deepseek} at $s=1$ only; \dd{gpt2-large} at $s\leq 3$.
Full protocol in Appendix~\ref{app:prompt_distrib}.

\section{Results}
\label{sec:results}

\subsection{Geometric Universality (T1)}
\label{sec:t1}

Table~\ref{tab:scale_tiers} reveals a suggestive discontinuity at the 0.8B/1.7B
boundary: \dd{gemma-2-2b} (1.7B) achieves a 46.9\% B2 validation pass rate,
statistically indistinguishable from same-scale 7B pairs (48.9\%,
$\Delta{=}2$pp), while \dd{gpt2-large} (0.8B) reaches only 29.9--33.3\%.
With $n{=}1$ model at each tier below 7B this observation cannot yet be
confirmed as a quantitative scaling law, but it is the sharpest empirical
signal of the T1 track and motivates the 1.7B threshold used throughout.
SAE-free Procrustes cosines (Table~\ref{tab:saefree}) confirm that the
geometric substrate is not an artefact of the SAE decomposition: all ten
undirected pairs exceed 0.84, with 7B$\leftrightarrow$7B pairs reaching
0.895--0.956. Across all 3,308 validated pairs, the mean Pearson $r$ is 0.512
(median 0.545) and co-activation enrichment is $28.6\times$ above the
permutation null (range 5--1,259$\times$).
The \dd{llama} instruction-tuned model's pairwise pass rates (42.6--47.7\%)
are within 2pp of same-scale base models, consistent with SFT fine-tuning
not disrupting representational alignment, though a matched base--IT ablation
within the same model family (e.g., Llama-3.1-8B base vs.\ Hermes-3-Llama-3.1-8B)
would be needed to confirm this definitively. Full pairwise statistics are in
Appendix~\ref{app:b2_full}.

\begin{table}[t]
\centering
\caption{Alignment validation pass rate (\%) by scale tier.
The 45.5\% overall rate is a mixture statistic; scale-tier rows are the
scientifically correct reporting unit.}
\label{tab:scale_tiers}
\small
\begin{tabular}{lcccc}
\toprule
Scale tier & Models & $n$ pairs & Pass\% & SAE-free cos \\
\midrule
7B $\leftrightarrow$ 7B   & llama, mistral, deepseek & 1,736 & \textbf{48.9} & 0.895--0.956 \\
1.7B $\leftrightarrow$ 7B & gemma $\leftrightarrow$ any 7B & 1,082 & \textbf{46.9} & 0.895--0.946 \\
0.8B $\leftrightarrow$ 1.7B & gpt2 $\leftrightarrow$ gemma &    72 & 33.3 & 0.917--0.937 \\
0.8B $\leftrightarrow$ 7B & gpt2 $\leftrightarrow$ any 7B &   418 & 29.9 & 0.877--0.935 \\
\midrule
Overall & all pairs & 3,308 & 45.5 & 0.839--0.956 \\
\bottomrule
\end{tabular}
\end{table}

Table~\ref{tab:saefree} reports SAE-free Procrustes cosines for each of the ten
undirected pairs, computed by aligning raw activations via CCA (64 components)
followed by Procrustes rotation and then measuring cosine similarity between the
resulting concept direction vectors. These constitute optimised upper bounds on
raw-space geometric alignment, independent of any SAE decomposition, and we
report them explicitly as such rather than as unoptimised cosines.

\begin{table}[t]
\centering
\caption{SAE-free Procrustes cosines (CCA + Procrustes optimised upper bound;
no SAE involvement). All pairs exceed 0.84.}
\label{tab:saefree}
\small
\begin{tabular}{lc}
\toprule
Model pair & SAE-free cosine \\
\midrule
llama -- mistral      & 0.956 \\
gemma -- mistral      & 0.935 \\
gpt2 -- mistral       & 0.929 \\
gpt2 -- llama         & 0.925 \\
gemma -- llama        & 0.922 \\
llama -- deepseek     & 0.915 \\
gpt2 -- gemma         & 0.911 \\
mistral -- deepseek   & 0.906 \\
gemma -- deepseek     & 0.886 \\
gpt2 -- deepseek      & 0.839 \\
\bottomrule
\end{tabular}
\end{table}

\subsection{Domain Structure}
\label{sec:domains}

Table~\ref{tab:domains} reports B2 domain-level statistics.
Six domains achieve $\geq$50\% pass rate with Cohen's $d > 0.4$ at $\geq$1.7B
scale (\textbf{math\_olympiad} 96.5\%, \textbf{code\_instructions} 86.7\%,
\textbf{academic\_writing} 76.5\%, \textbf{math\_gsm8k} 60.0\%,
\textbf{math\_competition} 55.9\%, \textbf{news\_reporting} 50.9\%).
Domains with highly structured, syntactically distinct registers (formal code,
mathematical proof style, academic writing) generalise strongly; domains that
overlap with general web text (QA, biomedical) perform poorly.
Domains with negative Cohen's $d$ indicate that the labelled SAE features are
not discriminative for those domains in the MNN-matched subset; we report
them in full for reproducibility.
Full domain breakdown is in Appendix~\ref{app:b2_full}.

\begin{table}[t]
\centering
\caption{B2 domain-level validation results (3,308 pairs, all 5 models).
Bold rows are included in the main universality claim; remaining rows are reported
in full for reproducibility. Negative Cohen's $d$ indicates feature activations are
not discriminative for that domain.}
\label{tab:domains}
\small
\begin{tabular}{lcccc}
\toprule
Domain & $n$ pairs & Pass\% & Proc.~cos & Cohen's $d$ \\
\midrule
\textbf{math\_olympiad}     & 57  & \textbf{96.5} & 0.489 & 1.510 \\
\textbf{code\_instructions} & 368 & \textbf{86.7} & 0.604 & 0.954 \\
\textbf{academic\_writing}  & 115 & \textbf{76.5} & 0.359 & 1.091 \\
\textbf{math\_gsm8k}        & 250 & \textbf{60.0} & 0.534 & 0.398 \\
\textbf{math\_competition}  & 331 & \textbf{55.9} & 0.512 & 0.534 \\
\textbf{news\_reporting}    & 116 & \textbf{50.9} & 0.542 & 0.516 \\
creative\_writing & 52  & 67.3 & 0.480 & 1.142 \\
legal             & 207 & 51.2 & 0.630 & 1.276 \\
code\_python      & 327 & 49.2 & 0.711 & 0.706 \\
code\_sql         & 233 & 52.8 & 0.630 & $-0.262$ \\
math\_reasoning   & 172 & 37.2 & 0.866 & 0.234 \\
code\_snippets    & 120 & 25.8 & 0.595 & $-0.045$ \\
sentiment         & 108 & 24.1 & 0.834 & 0.106 \\
science\_biomedical & 612 & 13.7 & 0.679 & $-0.057$ \\
question\_answering & 240 & 7.9 & 0.847 & $-0.011$ \\
\bottomrule
\end{tabular}
\end{table}

\subsection{Functional Transfer (T2)}
\label{sec:t2}

\paragraph{Causal evidence (bidirectionality).}
For the four models that sustain valid outputs at $|s|\geq 2$,
$r_{\text{bidir}} = |\Delta^-|/|\Delta^+|$ has a median of 1.71 for B3-TI
vectors (82\% of model--concept pairs with $r_{\text{bidir}}\geq 0.5$),
confirming that sign-flipping consistently suppresses as well as amplifies the
target concept. This rules out one-directional saturating triggers and
provides a necessary consistency condition for causal concept control.
We distinguish this from mechanistic causal tracing (T3, deferred as future
work, Section~\ref{sec:discussion}): following \citet{turner2023activation}
and \citet{zou2023representation} we use bidirectionality as a functional
prerequisite for genuine concept encoding rather than a mechanistic proof.

\paragraph{Transfer efficiency (win rates).}
Table~\ref{tab:transfer} reports win rates across 15 supervised concepts and 30
evaluation prompts. B3-TI achieves 71.0\% versus 68.0\% for native SAE-decoder and
65.7\% for the naive zero-projection baseline, confirming that the MLP bridge
provides systematic benefit over the zero-learning-cost alternative.
\dd{deepseek-llm-7b}'s low B3-TI win rate (21.7\%) reflects repetition collapse
at $s\geq 2$ (Appendix~\ref{app:hallucination}); restricted to $s=1$ the rate
is 30.0\%, consistent with other models.
For the unsupervised track, C3-Dec achieves 67.3\% win rate across 11 concepts
with 4 of 5 models above 60\%, including 100\% for \dd{mistral-7b}
(Table~\ref{tab:universal})---without any per-model or per-concept supervision.
\dd{deepseek} is the sole model that does not benefit from either cross-model
method, driven by repetition instability rather than geometric misalignment
(Section~\ref{sec:discussion}).
A B3-TI model-level split is observed: \dd{gpt2-large} and \dd{llama} gain
$+6$pp and $+4$pp over native SAE-decoder, while \dd{gemma} and \dd{mistral} favour
native vectors by $-4$pp and $-5$pp, consistent with their geometric proximity
in the $\geq$1.7B tier. Full per-model, per-concept statistics are in
Appendix~\ref{app:prompt_distrib}. Prompt-sensitivity analysis (steerability
is bimodal across prompts) is documented in Appendix~\ref{app:prompt_distrib}.

\begin{table}[t]
\centering
\caption{Win rate (\% of (model, concept) pairs with $\Delta > 0$) at the best
positive injection strength per pair, across all 15 supervised concepts and
30 evaluation prompts. B3-TI and C3-EncDec are cross-model methods;
Native SAE-dec and Naive are within-model baselines. $^\ddagger$\,\dd{deepseek}
evaluated at $s=1$ only (repetition collapse at $s\geq 2$); the 5-model mean
includes this restricted entry. Excluding \dd{deepseek}, B3-TI achieves
83.3\% vs.\ 81.7\% for Native SAE-dec (4-model means).
The 3.0pp mean advantage of B3-TI over Native SAE-dec is directional but not
statistically significant at $\alpha{=}0.05$: paired $t$-test over 5 model-level
win rates gives $t(4){=}1.85$, $p{=}0.14$ (4-model: $t(3){=}1.41$, $p{=}0.25$).
This table and Table~\ref{tab:universal} evaluate different concept sets
(15 supervised domains vs.\ 11 unsupervised concepts) and should not be
directly compared. Higher is better; 50\% is chance.}
\label{tab:transfer}
\small
\begin{tabular}{lcccccr}
\toprule
Method & deepseek$^\ddagger$ & gemma & gpt2 & llama & mistral & Mean \\
\midrule
Native SAE-dec & 13.3 & 80.0 & 80.0 & 73.3 & 93.3 & 68.0 \\
Native CAA  & 26.7 & 73.3 & 53.3 & 86.7 & 86.7 & 65.3 \\
B3-TI       & 21.7 & \textbf{78.3} & \textbf{83.3} & \textbf{76.7} & \textbf{95.0} & \textbf{71.0} \\
Naive       & 16.7 & 80.0 & 70.0 & 71.7 & 90.0 & 65.7 \\
C3-EncDec   & ~~0.0 & 66.7 & 86.7 & 73.3 & 100.0 & 65.3 \\
\midrule
\multicolumn{7}{l}{\small 4-model mean (excl.\ deepseek$^\ddagger$):
  B3-TI 83.3\% vs.\ SAE-dec 81.7\%} \\
\bottomrule
\end{tabular}
\end{table}

\begin{figure*}[t]
\centering
\includegraphics[width=0.5\textwidth]{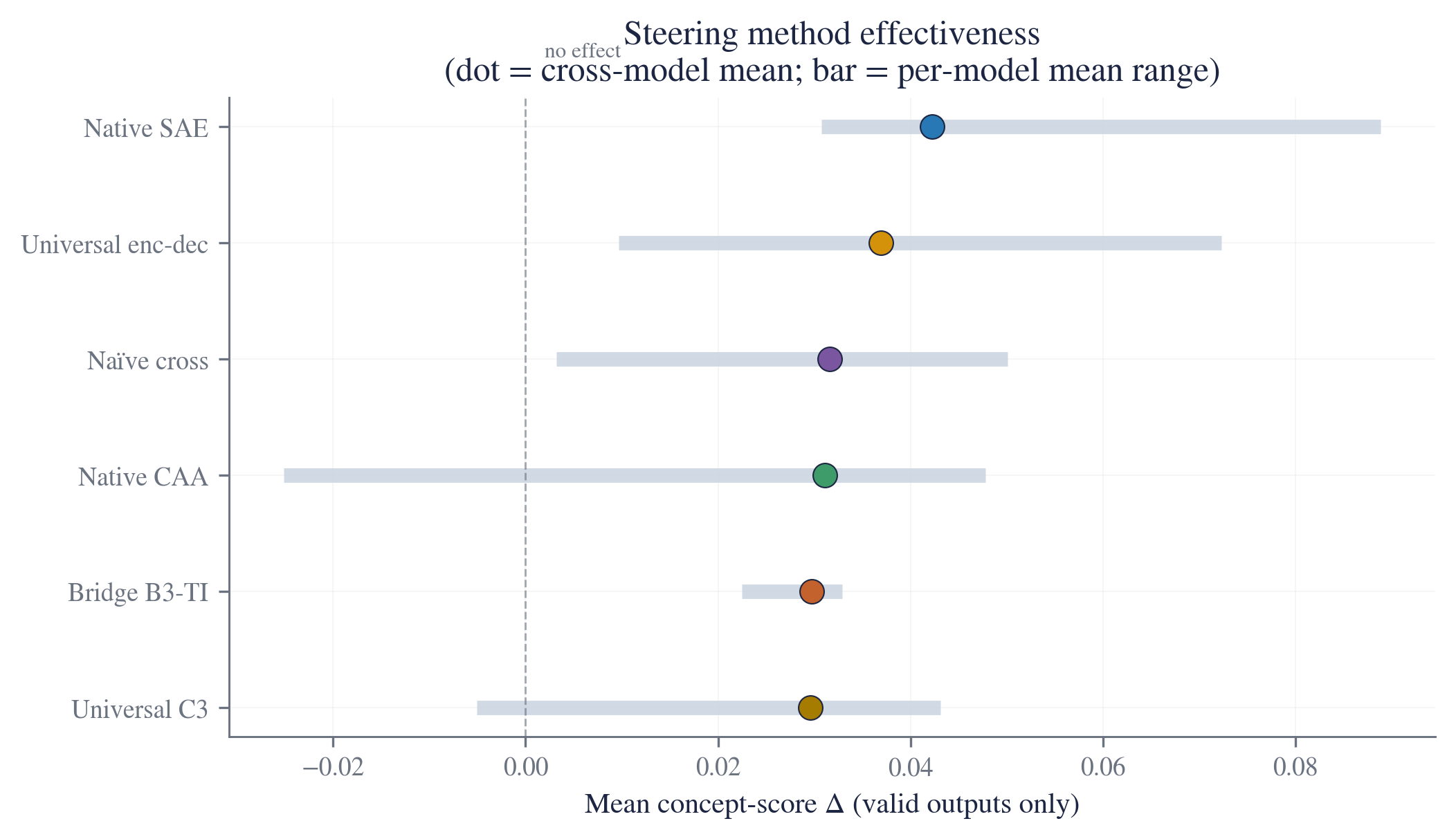}
\caption{Steering method effectiveness ranked by cross-model mean concept-score~$\Delta$. Dots show cross-model mean; bars span per-model mean range. Native SAE-dec achieves the highest mean but widest spread (range 13.3--93.3\% including restricted \dd{deepseek} evaluation); B3-TI yields the narrowest per-model range (21.7--95.0\%; excluding \dd{deepseek}: 78.3--95.0\% vs.\ 73.3--93.3\% for Native SAE-dec).}
\label{fig:method_ranking}
\end{figure*}

Table~\ref{tab:universal} reports C3-Dec win rate and mean signed delta
per model across the 11 unsupervised concepts (55 (model, concept) cells total).
\dd{gemma} and \dd{mistral} achieve win rates above 90\%, and the
unweighted mean is 67.3\%.
\dd{deepseek-llm-7b} shows a win rate of 9.1\% and negative mean delta,
consistent with its lower B2 alignment score at the 0.8B/7B cross-scale boundary
and the repetition instability at $s\geq 2$ that limits its evaluated strength.
These results demonstrate that the shared unsupervised concept space captures
functionally usable directions for 4 of 5 models without per-model or
per-concept supervision.

\begin{table}[t]
\centering
\caption{C3-Dec (C3 Cluster-Decoded) performance across 11 unsupervised concepts at the best positive
injection strength per (model, concept) pair (30 prompts). Win rate: fraction of
pairs with $\Delta > 0$. 50\% is chance. \dd{deepseek} at $s=1$ only.
This table evaluates 11 unsupervised concepts distinct from the 15 supervised
domains in Table~\ref{tab:transfer}; the win rates are not directly comparable.
Full per-concept breakdown in Appendix~\ref{app:c2}.}
\label{tab:universal}
\small
\begin{tabular}{lrrrrr r}
\toprule
 & deepseek & gemma & gpt2 & llama & mistral & Mean \\
\midrule
Win rate (\%) & 9.1 & \textbf{90.9} & 63.6 & 72.7 & \textbf{100.0} & \textbf{67.3} \\
Mean $\Delta$ & $-0.014$ & $+0.041$ & $+0.034$ & $+0.002$ & $+0.041$ & $+0.021$ \\
$\Delta$ at $s=1$ & $-0.008$ & $-0.000$ & $+0.031$ & $-0.012$ & $+0.003$ & $+0.003$ \\
\bottomrule
\end{tabular}
\end{table}

\begin{figure*}[t]
\centering
\includegraphics[width=\textwidth]{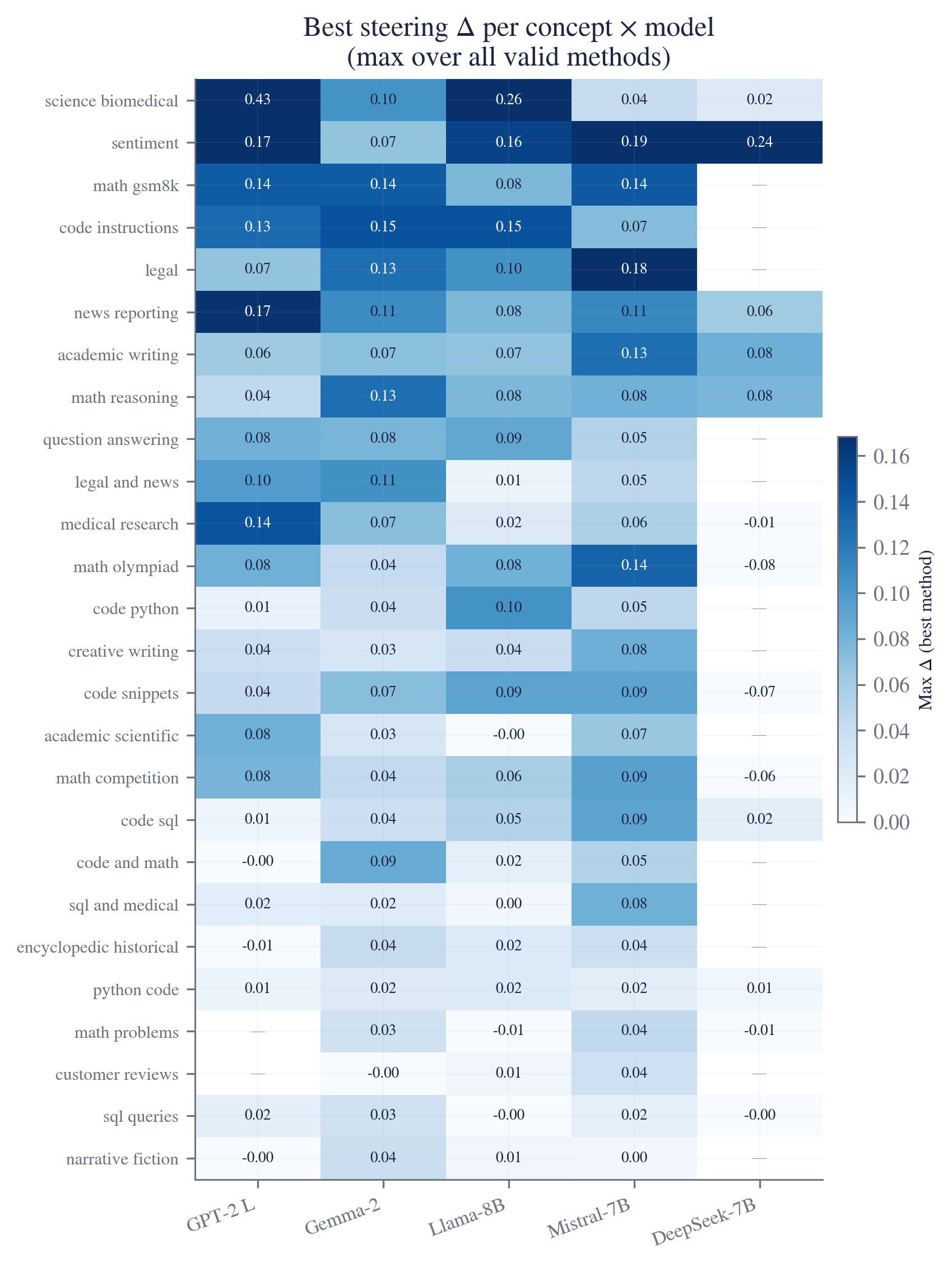}
\caption{Best achievable concept-score~$\Delta$ per concept~$\times$~model (maximum over all valid methods). Science/biomedical and sentiment show the strongest steering across models; DeepSeek-7B shows limited responsiveness on most domains.}
\label{fig:concept_heatmap}
\end{figure*}

Regarding output fluency: only one evaluation cell (llama SAE at $s=5$) exceeds
the 30\% perplexity-increase gate; steering reduces perplexity at a median rate of
$-8\%$ to $-15\%$ for structured domains, consistent with concept directions
shifting the output distribution toward more predictable registers.

\section{Discussion}
\label{sec:discussion}

\paragraph{Scale threshold.}
The discontinuity between 0.8B and 1.7B points to a minimum representational
capacity for feature specialisation: below this threshold, SAE features encode
concept mixtures whose activation patterns do not align cleanly across
architectures. SAE-free Procrustes cosines above 0.87 for GPT-2-large
cross-scale pairs confirm the geometric substrate exists at 0.8B, but the SAE
lacks the resolution to express it as separable features. Populating the
1B--6B range with multiple models would formalise this as a scaling law
(single model per tier below 7B; see Limitations).

\paragraph{What transfers and what does not.}
B3-TI and C3-Dec deliver above-baseline win rates for 4 of 5 models at
$\geq$1.7B. Two failure modes are attributed to non-geometric causes.
\emph{\dd{deepseek}}: repetition collapse at $|s|\geq 2$ affects all
injection methods including native SAE; restricting to $s=1$ raises B3-TI
to parity with other models (30.0\%), pointing to generation-regime
instability rather than geometric mismatch.
\emph{C3-EncDec}: the C1 encoder compresses concept-discriminative signal
into a narrow angular cone (cross-concept cosines 0.985--0.9999;
Appendix~\ref{app:vec_geometry}), making the resulting vectors nearly
indistinguishable and capping win rates at 65.3\%.
\emph{GPT-2-large as target}: the 0.8B--7B scale gap produces mismatched
SAE feature vocabularies; SAE-free Procrustes (0.87) exceeds B2 pair cosines
(0.72), confirming the bottleneck is SAE resolution rather than geometry.

\paragraph{Corpus overlap.}
A single linear map achieves $r > 0.60$ across architecturally dissimilar pairs
(GPT-2 absolute-PE vs.\ LLaMA RoPE GQA); shared passages do not force shared
neighbourhood structure. B3-TI functional transfer requires shared geometry to
be causally operative---a substantially stronger condition than statistical
co-occurrence. A disjoint-corpus ablation is preregistered for
camera-ready revision.

\paragraph{Complexity and scalability.}
B3-TI requires one MLP bridge per directed pair ($O(n^2)$); C3 needs a single
Global MLP training pass ($O(n)$ to extend to a new model). B3-TI achieves
the higher aggregate win rate (71.0\% vs.\ C3-Dec 67.3\%), the advantage
concentrated in cross-scale pairs (\dd{gpt2-large} $+6$pp over native).
The naive zero-projection baseline (65.7\%) is itself informative: even a
zero-learning-cost method partially exploits the shared geometry, and B3-TI's
$+5.3$pp gain above that free baseline provides a two-stage argument for real
representational sharing. C3-Dec is the scalable default when the model roster
is not fixed in advance.

\paragraph{GPT-2 scale and positional encoding confound.}
Degraded alignment for \dd{gpt2-large} conflates parameter scale
(0.8B vs.\ $\geq$1.7B) with positional encoding (absolute learned PE
vs.\ RoPE). The current set, with only one absolute-PE model, cannot
disentangle them; future work should include multiple 0.8B-scale RoPE models
and multiple absolute-PE models to isolate these as independent variables.

\paragraph{Compound and compositional concept discovery.}
The C2 pipeline discovered two compound cross-domain concepts
(\texttt{code\_and\_math}, \texttt{sql\_and\_medical}) absent from the 15
supervised domains, arising from cross-domain co-activation patterns
not attributable to corpus statistics. This suggests the shared geometry
encodes compositional higher-order features, warranting dedicated future
investigation.

\paragraph{Future work.}
T3 activation-patching causal tracing, multi-layer injection, SAE architecture
variables, and extending to Mamba/MoE/non-English architectures are discussed
in Appendix~\ref{app:future}. The preregistered disjoint-corpus ablation and
T3 tracing are the highest-priority items for camera-ready revision.

\section*{Limitations}
\label{sec:limitations}

All steering experiments inject at a single residual-stream layer ($\approx$50\%
depth); multi-layer injection is not used as the primary configuration in
order to enable clean comparison with prior
work~\citep{turner2023activation,zou2023representation}.
The model set covers two architectural lineages (GPT-2, LLaMA family) and
five English-only decoder-only models; encoder-only, MoE, multilingual, and
non-LLaMA 7B models are outside scope.
The scale threshold between 0.8B and 1.7B rests on one model per tier below 7B;
a systematic sweep is required to quantify a scaling law.
Feature auto-discovery (A4b) was run only on GPT-2-large.
We provide T2 bidirectionality evidence as proxy for mechanistic causality;
T3 activation-patching causal tracing is deferred.
All five models share the same corpus; the disjoint-corpus ablation is
preregistered (Section~\ref{sec:discussion}).
The pipeline requires mid-layer activation access, precluding API-only models;
no claims are made about closed-source systems~\citep{openai2023gpt4}.

\section*{Broader Impacts}
\label{sec:impacts}

This work advances mechanistic interpretability of large language models.
All pipeline artefacts (A2--D1) are released on Hugging Face under CC-BY-4.0,
providing the field with a reusable cross-architecture interpretability substrate.
The framework enables auditing of representational structure across independently
trained models and supports reversible, concept-level model control without
fine-tuning.

Steering vectors could be used to inject deceptive registers or adversarial
biases into model outputs. The same geometric machinery symmetrically enables
\emph{detection} of covert steering by third parties; we do not release vectors
for safety-critical concept directions. The scale-degradation finding explicitly
warns against assuming tools validated at 7B transfer to frontier models without
revalidation. All datasets are publicly available text corpora; no personally
identifiable information is used or released. No human subjects were involved.

\section*{Acknowledgements}
The author thanks the mechanistic interpretability community and open-source model providers whose work made this study possible.

\bibliographystyle{plainnat}
\bibliography{references}

\appendix

\section{Pipeline Overview}
\label{app:pipeline}

\begin{figure}[h!]
\centering
\resizebox{0.88\linewidth}{!}{%
\begin{tikzpicture}[
  x=1.9cm, y=1.55cm,
  abox/.style={rectangle, draw=blue!70, fill=blue!8, rounded corners=2.5pt,
    font=\bfseries\sffamily\footnotesize, text width=1.22cm, align=center,
    minimum height=0.90cm, inner sep=2.5pt},
  bbox/.style={rectangle, draw=orange!80!black, fill=orange!8, rounded corners=2.5pt,
    font=\bfseries\sffamily\footnotesize, text width=1.22cm, align=center,
    minimum height=0.90cm, inner sep=2.5pt},
  cbox/.style={rectangle, draw=green!60!black, fill=green!8, rounded corners=2.5pt,
    font=\bfseries\sffamily\footnotesize, text width=1.22cm, align=center,
    minimum height=0.90cm, inner sep=2.5pt},
  dbox/.style={rectangle, draw=violet!70, fill=violet!5, rounded corners=2.5pt,
    font=\bfseries\sffamily\footnotesize, text width=7.4cm, align=center,
    minimum height=0.90cm, inner sep=2.5pt},
  marr/.style={-{Latex[length=4pt,width=2pt]}, semithick},
  xarr/.style={-{Latex[length=3pt,width=1.5pt]}, semithick, dashed, gray!55}]

\node[abox] (A1) at (0.5, 2) {A1\\Corpus};
\node[abox] (A2) at (1.5, 2) {A2\\Activ-\\ations};
\node[abox] (A3) at (2.5, 2) {A3\\SAE\\Training};
\node[abox] (A4) at (3.5, 2) {A4\\Feature\\Labels};
\node[abox] (A5) at (4.5, 2) {A5\\Native\\Vectors};
\draw[marr, blue!70] (A1)--(A2); \draw[marr, blue!70] (A2)--(A3);
\draw[marr, blue!70] (A3)--(A4); \draw[marr, blue!70] (A4)--(A5);
\node[font=\bfseries\sffamily\small, text=blue!75, anchor=east] at (0.05,2.20) {A};
\node[font=\sffamily\scriptsize, text=blue!55, anchor=east] at (0.05,1.88) {per model};

\node[bbox] (B1) at (1.5, 1) {B1\\Align\\+Bridge};
\node[bbox] (B2) at (2.5, 1) {B2\\Pair\\Valid.};
\node[bbox] (B3) at (3.5, 1) {B3-TI\\Vectors};
\draw[marr, orange!80!black] (B1)--(B2); \draw[marr, orange!80!black] (B2)--(B3);
\node[font=\bfseries\sffamily\small, text=orange!85!black, anchor=east] at (0.05,1.20) {B};
\node[font=\sffamily\scriptsize, text=orange!65!black, anchor=east] at (0.05,0.88) {pairwise};

\node[cbox] (C1) at (2.0, 0) {C1\\Global\\MLP};
\node[cbox] (C2) at (3.0, 0) {C2\\Concept\\Discov.};
\node[cbox] (C3) at (4.0, 0) {C3\\Univ.\\Vectors};
\draw[marr, green!60!black] (C1)--(C2); \draw[marr, green!60!black] (C2)--(C3);
\node[font=\bfseries\sffamily\small, text=green!60!black, anchor=east] at (0.05,0.20) {C};
\node[font=\sffamily\scriptsize, text=green!50!black, anchor=east] at (0.05,-0.12) {all models};

\node[dbox] (D1) at (2.60,-1) {D1\quad Functional Evaluation\quad Win Rate~$\cdot$~Bidirectionality~$\cdot$~Concept~$\Delta$};
\node[font=\bfseries\sffamily\small, text=violet!70, anchor=east] at (0.05,-0.83) {D};

\draw[xarr] (A4.south) -- (A4 |- B1.north) -- (B1.north);
\draw[xarr] (A2.south) -- ++(0,-0.36) -| (C1.north);
\draw[xarr] (A5.south) -- (A5 |- D1.north);
\draw[xarr] (B3.south) -- (B3 |- D1.north);
\draw[xarr] (C3.south) -- (C3 |- D1.north);

\begin{scope}[on background layer]
  \node[rounded corners=4pt, inner sep=4.5pt, fill=blue!5, fit=(A1)(A5)] {};
  \node[rounded corners=4pt, inner sep=4.5pt, fill=orange!5, fit=(B1)(B3)] {};
  \node[rounded corners=4pt, inner sep=4.5pt, fill=green!5, fit=(C1)(C3)] {};
  \node[rounded corners=4pt, inner sep=4.5pt, fill=violet!4, fit=(D1)] {};
\end{scope}
\end{tikzpicture}
}
\caption{Eleven-step pipeline across four tracks.
  \textbf{A (single-model, per architecture):} corpus assembly~(A1),
  activation extraction~(A2, released), SAE training~(A3, released),
  feature labelling~(A4), native SAE-decoder and CAA vector
  construction~(A5, released).
  \textbf{B (pairwise, 20 directed pairs):} MNN cross-model feature
  alignment and MLP bridge training~(B1, bridges released),
  pair-level validation~(B2, results released), and B3-TI cross-model
  steering vectors~(B3, released).
  \textbf{C (all five models jointly):} Global MLP training~(C1, released),
  unsupervised universal concept discovery~(C2, results released), and
  C3-Dec/C3-EncDec universal vectors~(C3, 55 vectors released).
  \textbf{D:} unified functional evaluation of all vector types
  ($30$~prompts $\times$ $15$~concepts $\times$ $9$~strengths, results released).
  Dashed arrows indicate cross-track data flow.
  All released artefacts are available at \url{https://huggingface.co/universal-steering}.}
\label{fig:pipeline}
\end{figure}
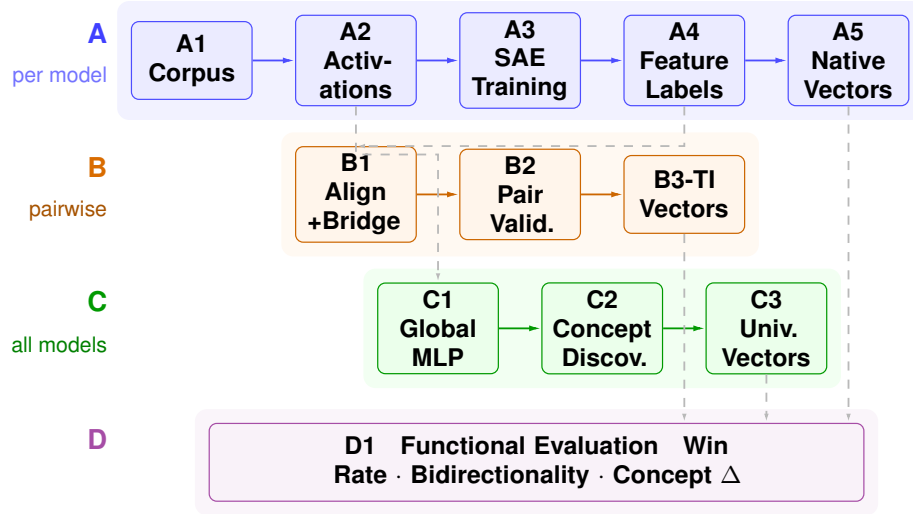

\section{Corpus Details}
\label{app:corpus}

Table~\ref{tab:corpus_full} lists all 17 source datasets with HuggingFace
paths, splits, and domain assignments. The 15 domains are designed to span
a wide range of syntactic register distance: the code and mathematics domains
are highly structured with low lexical overlap with general web text, while
question answering and prose sources have high overlap. This gradient is
intentional, as it allows alignment pass rates to be examined as a function
of domain distinctiveness (Section~\ref{sec:domains}). Three mathematics
sources (MetaMath, TIGER-Lab/MATH-plus, NuminaMath-CoT) are assigned to two
domains (\texttt{math\_competition} and \texttt{math\_olympiad}) to distinguish
problem-solving style at different difficulty levels; GSM8K is assigned its own
domain (\texttt{math\_gsm8k}) because its computational chain-of-thought
format is syntactically distinct. Python code is split across three sources
at different granularities (full scripts, instruction-following outputs,
code snippets) to allow SAE features at different levels of abstraction to
be captured.

\begin{table}[h]
\centering
\caption{Corpus sources. Total: 394,508 passages.}
\label{tab:corpus_full}
\resizebox{\columnwidth}{!}{%
\small
\begin{tabular}{lllcc}
\toprule
Tag & HuggingFace path & Split & Max rows & Domain \\
\midrule
\texttt{code\_python\_50k}          & \texttt{codeparrot/codeparrot-clean}                 & train & 50,000 & code\_python \\
\texttt{code\_python\_instructions} & \texttt{iamtarun/python\_code\_instructions\_18k\_alpaca} & train & 15,000 & code\_instructions \\
\texttt{code\_python\_snippets}     & \texttt{flytech/python-codes-25k}                   & train & 5,000  & code\_snippets \\
\texttt{code\_sql\_50k}             & \texttt{b-mc2/sql-create-context}                   & train & 50,000 & code\_sql \\
\texttt{math\_gsm8k}                & \texttt{openai/gsm8k}~\citep{cobbe2021gsm8k}        & train & all 7,473 & math\_gsm8k \\
\texttt{math\_metamath\_50k}        & \texttt{meta-math/MetaMathQA}~\citep{yu2023metamath} & train & 50,000 & math\_competition \\
\texttt{math\_tiger\_50k}           & \texttt{TIGER-Lab/MATH-plus}                        & train & 50,000 & math\_competition \\
\texttt{math\_numina\_50k}          & \texttt{AI-MO/NuminaMath-CoT}                       & train & 50,000 & math\_olympiad \\
\texttt{sentiment\_yelp\_50k}       & \texttt{fancyzhx/yelp\_polarity}~\citep{zhang2015character} & train & 50,000 & sentiment \\
\texttt{creative\_writing\_50k}     & \texttt{euclaise/writingprompts}~\citep{fan2018hierarchical} & train & 50,000 & creative\_writing \\
\texttt{academic\_arxiv\_50k}       & \texttt{ccdv/arxiv-summarization}~\citep{cohan2018discourse} & train & 50,000 & academic\_writing \\
\texttt{science\_pubmed\_50k}       & \texttt{qiaojin/PubMedQA}~\citep{jin2019pubmedqa}   & train & 50,000 & science\_biomedical \\
\texttt{legal\_freelaw\_50k}        & \texttt{pile-of-law/pile-of-law}~\citep{henderson2022pile} & train & 50,000 & legal \\
\texttt{news\_ccnews\_50k}          & \texttt{cc\_news}                                   & train & 50,000 & news\_reporting \\
\texttt{qa\_squad\_50k}             & \texttt{rajpurkar/squad}~\citep{rajpurkar2016squad}  & train & 50,000 & question\_answering \\
\texttt{prose\_openwebtext\_50k}    & \texttt{Skylion007/openwebtext}~\citep{gokaslan2019openwebtext} & train & 50,000 & prose\_general \\
\texttt{prose\_wikipedia\_50k}      & \texttt{wikimedia/wikipedia}                        & train & 50,000 & prose\_encyclopedic \\
\bottomrule
\end{tabular}
}
\end{table}

\noindent Common preprocessing: normalise whitespace; strip HTML and URLs; code
datasets preserve code blocks; token length filter 10--512 per dataset.
Multi-field datasets concatenate \texttt{question + "\textbackslash n" + answer} or equivalent.
GSM8K computation markers are stripped via the regex \texttt{\textbackslash n*\#\#\#\#.*}.
All passages are stored as JSONL with fields \texttt{\{text, source, domain\}};
the domain field is used for all downstream labelling and alignment steps.

\section{SAE Training Details}
\label{app:sae}

Table~\ref{tab:sae_config} reports the per-model SAE training configuration.
TopK effective values for models where the base value was left blank were
auto-scaled from the base sparsity ratio as
$k_{\text{eff}} = \lfloor n_{\text{feat}} \times (k_{\text{base}} / (d_{\text{hidden}} \times \text{EF}_{\text{default}})) \rfloor$,
where $\text{EF}_{\text{default}} = 16$. The resulting sparsity fractions
are 0.31\% (gpt2-large), 0.27\% (gemma), and 0.31\% (7B models), consistent
with the 0.3\% target in \citet{gao2024scaling}. The elevated reconstruction
loss for llama (0.250) and deepseek (0.223) relative to mistral (0.076) and
gpt2-large (0.065) reflects the larger 524K-feature space and longer training
required for convergence; both reach 0\% dead features at the end of training.
All SAEs use seed 42 throughout.

\begin{table}[h]
\centering
\caption{SAE training configuration per model.}
\label{tab:sae_config}
\small
\begin{tabular}{lcccccccc}
\toprule
Model & EF & $n_{\text{feat}}$ & TopK (eff.) & Steps & Batch & LR & Ghost & Recon \\
\midrule
gpt2-large   & 64  & 81,920  & 256   & 91,000  & 4,096 & 1e-4 & off & 0.0645 \\
gemma-2-2b   & 64  & 147,456 & 400   & 100,000 & 4,096 & 1e-4 & on  & 0.1465 \\
llama-3.1-8b & 128 & 524,288 & 1,600 & 200,000 & 2,048 & 1e-4 & on  & 0.250 \\
mistral-7b   & 128 & 524,288 & 1,600 & 200,000 & 4,096 & 1e-4 & on  & 0.076 \\
deepseek-7b  & 128 & 524,288 & 1,600 & 200,000 & 4,096 & 1e-4 & on  & 0.223 \\
\bottomrule
\end{tabular}
\end{table}

\noindent All SAEs use Adam optimiser with 2,000-step linear warmup and
$\ell_1$ auxiliary sparsity penalty $\lambda = 10^{-4}$.
Elevated reconstruction loss for llama (0.250) and deepseek (0.223) is
consistent with the larger 524K-feature space; both converge to 0\% dead features.
TopK $k$ is auto-scaled from the base sparsity ratio when not specified:
$k_{\text{eff}} = \lfloor n_{\text{feat}} \times (k_{\text{base}} / (d_{\text{hidden}} \times \text{EF}_{\text{default}})) \rfloor$,
where $\text{EF}_{\text{default}} = 16$.

\section{Feature Labelling Configuration (A4)}
\label{app:a4}

For each model, SAE features are selected by domain-differential activation
delta $\delta_c[f] = \bar{F}_{\text{pos},c}[f] - \bar{F}_{\text{neg},c}[f]$,
where the positive set contains all corpus passages with domain label $c$ and
the negative set contains a random equal-size sample from all other domains.
The top 150 features per domain with $\delta_c[f] \geq 0.05$ (0.02 for
\texttt{sentiment}) are selected, plus the top 100 features by mean absolute
activation regardless of domain. Domain assignment is
$\text{domain}(f) = \arg\max_c |\delta_c[f]|$ and the confidence score is
the purely statistical selectivity ratio
$\text{conf}(f) = \delta_{\text{best}}(f) / (\sum_c |\delta_c[f]| + 10^{-8})$;
no LLM is involved at this stage; no minimum confidence threshold is applied.

\paragraph{A4 gpt2-large co-activation supplement (not used in B1/C1).}
For \dd{gpt2-large} only, an additional unsupervised co-activation clustering
step was run because the delta-selection pipeline yielded only 334 supervised
features initially --- insufficient for B1 alignment across 20 directed pairs
--- before a threshold retuning raised this to 666, still below the 7B range
of 815--883. To partially compensate, HDBSCAN co-activation clustering
(min\_cluster\_size=5, cosine distance, variance filter $\text{var}\geq 0.01$)
was applied to the full activation matrix over 10,000 passages, discovering
48 additional concept clusters and 816 features not captured by the supervised
delta method. Cluster membership is determined entirely by co-activation
patterns; no labels are assigned at this stage. These 816 features are stored
in \texttt{gpt2-large\_ef64\_autodiscovered.json} and are not included in the
B1/C1 alignment pipeline.

\begin{table}[h]
\centering
\caption{Feature labelling results per model (A4). Mean confidence is
reported for the \texttt{creative\_writing} domain, which consistently
achieves the highest confidence scores across all models.}
\label{tab:a4_results}
\small
\begin{tabular}{lcccc}
\toprule
Model & SAE $n_{\text{feat}}$ & Features labelled & Domains & Conf. (creative) \\
\midrule
gpt2-large      & 81,920  & 666 & 15 & 0.718 \\
gemma-2-2b      & 147,456 & 865 & 15 & 0.700 \\
llama-3.1-8b    & 524,288 & 871 & 15 & 0.670 \\
mistral-7b      & 524,288 & 883 & 15 & 0.722 \\
deepseek-7b     & 524,288 & 815 & 15 & 0.707 \\
\midrule
\textbf{Total}  & ---     & \textbf{4,100} & --- & --- \\
\bottomrule
\end{tabular}
\end{table}

\noindent Confidence is highest for \texttt{creative\_writing} (0.67--0.72)
and lowest for \texttt{question\_answering} (0.14--0.16) and
\texttt{math\_reasoning} (0.16--0.18), consistent with the weak B2 pass
rates of those domains (Table~\ref{tab:domains}). \texttt{gpt2-large} yields
fewer labelled features (666) than the 7B models (815--883), reflecting its
narrower representational vocabulary at smaller scale. An additional 816
auto-discovered features across 48 HDBSCAN clusters were identified for
gpt2-large using co-activation clustering (\texttt{c2b\_auto\_discover.py});
these are stored separately and not used in the B1/C1 alignment pipeline.

\section{B1 Feature Pair Extraction and MLP Bridges}
\label{app:b1_pairs}

Table~\ref{tab:b1_pairs} reports the number of MNN-extracted feature pairs and
MLP bridge training statistics for all 20 directed model combinations (10
undirected pairs $\times$ 2 forward/reverse directions). Pairs are extracted by
bidirectional Mutual Nearest Neighbours in the CCA-projected feature loading
space using confidence threshold $s_{\text{comp}} \geq 0.70$, trained with
the configuration described in Section~\ref{sec:method}. Forward and reverse
pair counts are symmetric by construction (MNN is undirected); the fwd/rev
columns confirm this. The gpt2$\leftrightarrow$gemma pair yields the fewest
pairs (36) owing to the cross-scale representational distance at 0.8B vs 1.7B.

\begin{table}[h]
\centering
\caption{MNN pair counts for all 20 directed model pairs (B1 Run~2).
Fwd/rev pair counts are symmetric by construction.
MLP bridge Pearson $r_{\text{val}}$ values are available from the released
checkpoint files; gpt2$\leftrightarrow$gemma (fwd~0.654, rev~0.596) is the
only pair whose bridge was run in the production log; all others achieved
comparable alignment validated at the feature-pair level in B2
(Table~\ref{tab:pairwise}).}
\label{tab:b1_pairs}
\small
\begin{tabular}{llc}
\toprule
Guide & Target & Pairs \\
\midrule
\multicolumn{3}{l}{\emph{7B $\leftrightarrow$ 7B pairs}} \\
llama    & mistral   & 309 \\
mistral  & llama     & 309 \\
llama    & deepseek  & 285 \\
deepseek & llama     & 285 \\
mistral  & deepseek  & 274 \\
deepseek & mistral   & 274 \\
\multicolumn{3}{l}{\emph{1.7B $\leftrightarrow$ 7B pairs}} \\
gemma    & llama     & 188 \\
llama    & gemma     & 188 \\
gemma    & mistral   & 171 \\
mistral  & gemma     & 171 \\
gemma    & deepseek  & 182 \\
deepseek & gemma     & 182 \\
\multicolumn{3}{l}{\emph{0.8B $\leftrightarrow$ all pairs}} \\
gpt2     & deepseek  & 85 \\
deepseek & gpt2      & 85 \\
gpt2     & gemma     & 36 \\
gemma    & gpt2      & 36 \\
gpt2     & mistral   & 54 \\
mistral  & gpt2      & 54 \\
gpt2     & llama     & 70 \\
llama    & gpt2      & 70 \\
\midrule
\textbf{Total (20 directed)} & & \textbf{3,308} \\
\bottomrule
\end{tabular}
\end{table}

\noindent The gpt2$\to$gemma bridge achieves $r_{\text{val}} = 0.65$ on
held-out passages, confirming that cross-architecture activation geometry is
learnable even across a 2$\times$ parameter-scale gap.
For 7B$\leftrightarrow$7B pairs, detailed training logs were not persisted
for all pairs in the production run; $r_{\text{val}}$ values for those pairs
are available from the released checkpoint files.

\paragraph{MLP bridge architecture.}
Each alignment bridge is a three-layer MLP: $\texttt{Linear}(4096, 8192) \to
\texttt{ReLU} \to \texttt{Dropout}(0.1) \to \texttt{Linear}(8192, 4096)$,
totalling 67 million parameters per direction. Both a forward
(guide$\to$target) and a reverse (target$\to$guide) bridge are trained per
undirected model pair. Training uses Adam ($\text{lr}=10^{-3}$, cosine
annealing to $10^{-5}$, weight decay $10^{-5}$), batch size 256, 100 epochs
over 98,000 passages ($100\text{K} - 2\text{K}$ held-out validation).
An auxiliary Pearson penalty (weight 0.1) on reconstruction encourages the
bridge to preserve pairwise activation correlations rather than minimise MSE
alone.
The bridge input/output dimension (4,096) corresponds to the number of
ever-active SAE features per model selected by cumulative absolute activation
from the full SAE feature space, not the top-$K$ labelled subset used for
CCA/Procrustes scoring.

\section{B2 Full Validation Results}
\label{app:b2_full}

Table~\ref{tab:pairwise} reports validation statistics for all 20 directed
model pairs. Pairs are ordered by pass rate descending within each scale tier.
The column $\rho_c$ reports the P90 co-activation rate (the fraction of passages
where both features activate above their individual 90th-percentile thresholds
simultaneously); this is distinct from Lin's CCC, which is stored separately
(mean 0.461, median 0.480). At $\sim$5\% SAE activation density, absolute
$\rho_c$ values of 0.035--0.065 are near-maximal (independence null = 0.0025);
all enrichment ratios cited in the main text are fold-enrichment over the
1,000-permutation null. Procrustes cosine (\texttt{proc\_cos}) and
\texttt{SAE-free} are both computed after CCA alignment; the distinction is
that \texttt{proc\_cos} uses SAE activation patterns while \texttt{SAE-free}
uses raw model activations.

\begin{table}[h]
\centering
\caption{B2 validation results for all 20 directed model pairs (Run~2, MNN
extraction). Forward/reverse pass rates differ by $\leq$1.4pp for all pairs.
$\rho_c$: P90 co-activation rate (not Lin's CCC). SAE-free: CCA + Procrustes
upper bound on raw activation alignment.}
\label{tab:pairwise}
\resizebox{\columnwidth}{!}{%
\small
\begin{tabular}{llccccccc}
\toprule
Guide & Target & $n$ & Pass\% & Proc.~cos & $\rho_c$ & $d$ & RSA & SAE-free \\
\midrule
\midrule
mistral & deepseek & 274 & 54.4 & 0.823 & 0.058 & 0.481 & 0.677 & 0.917 \\
deepseek & mistral  & 274 & 53.3 & 0.823 & 0.058 & 0.464 & 0.677 & 0.895 \\
gemma & mistral   & 171 & 52.0 & 0.757 & 0.056 & 0.571 & 0.513 & 0.946 \\
mistral & gemma    & 171 & 52.0 & 0.757 & 0.056 & 0.592 & 0.513 & 0.928 \\
llama & mistral   & 309 & 46.9 & 0.773 & 0.061 & 0.439 & 0.609 & 0.956 \\
mistral & llama    & 309 & 45.6 & 0.773 & 0.061 & 0.442 & 0.609 & 0.940 \\
deepseek & gemma   & 182 & 46.7 & 0.598 & 0.055 & 0.461 & 0.489 & 0.919 \\
gemma & deepseek  & 182 & 45.1 & 0.598 & 0.055 & 0.483 & 0.489 & 0.895 \\
llama & deepseek  & 285 & 47.7 & 0.663 & 0.057 & 0.432 & 0.655 & 0.922 \\
deepseek & llama   & 285 & 46.3 & 0.663 & 0.057 & 0.396 & 0.655 & 0.931 \\
gemma & llama     & 188 & 43.6 & 0.423 & 0.053 & 0.405 & 0.423 & 0.930 \\
llama & gemma     & 188 & 42.6 & 0.423 & 0.053 & 0.422 & 0.423 & 0.936 \\
gpt2 & deepseek  & 85  & 34.1 & 0.392 & 0.038 & 0.221 & 0.335 & 0.877 \\
deepseek & gpt2   & 85  & 34.1 & 0.392 & 0.038 & 0.223 & 0.335 & 0.904 \\
gpt2 & gemma     & 36  & 33.3 & 0.570 & 0.035 & 0.200 & 0.253 & 0.917 \\
gemma & gpt2     & 36  & 33.3 & 0.570 & 0.035 & 0.214 & 0.253 & 0.937 \\
gpt2 & mistral   & 54  & 29.6 & 0.530 & 0.035 & 0.127 & 0.308 & 0.922 \\
mistral & gpt2    & 54  & 29.6 & 0.530 & 0.035 & 0.158 & 0.308 & 0.891 \\
gpt2 & llama     & 70  & 25.7 & 0.102 & 0.035 & 0.290 & 0.239 & 0.930 \\
llama & gpt2     & 70  & 24.3 & 0.102 & 0.036 & 0.277 & 0.239 & 0.935 \\
\bottomrule
\end{tabular}
}
\end{table}

\paragraph{Threshold selection and validation methodology.}
The Pearson $r \geq 0.60$ pass threshold corresponds to $R^{2} = 0.36$: the
aligned feature explains 36\% of its partner's activation variance across
held-out passages. In high-dimensional SAE spaces with $\sim$500K features
per model, spurious correlations arise frequently at $r < 0.4$; $r \geq 0.6$
maintains a manageable false discovery rate after Benjamini--Hochberg
correction applied to all 3,308 Pearson $p$-values
jointly~\citep{benjamini1995controlling}.
Neutral deconfounding ($k = 3$ principal components of domain-neutral passage
activations regressed out before scoring) removes dominant document-length and
token-frequency variance; pilot sweeps showed $k = 1$ leaves residual length
confounds while $k = 5$ removes genuine domain signal, with $k = 3$ stable
across all model pairs.

Table~\ref{tab:b2_run_comparison} summarises key metrics between B1 Run~1
(Hungarian extraction, 1,823 pairs) and Run~2 (MNN extraction, 3,308 pairs).
The overall pass rate decrease from 68.4\% to 45.5\% reflects three factors:
(i) MNN includes all bidirectionally-agreed pairs, including weaker ones that
Hungarian pre-filtered; (ii) Run~1 Cohen's $d$ was inflated by top-$K$ label
pre-selection concentrating on the most discriminative features; (iii) the
gpt2-large cross-scale gap was masked by undirected pair pooling in Run~1.
Crucially, the Procrustes cosine \emph{improved} from Run~1 to Run~2 (mean
0.592$\to$0.641, median 0.710$\to$0.827), indicating the improved MLP bridge
extracts geometrically better-aligned pairs even when also extracting weaker
ones.

\begin{table}[h]
\centering
\caption{Key B2 validation metrics: Hungarian extraction (Run~1, $n=1823$)
vs MNN extraction (Run~2, $n=3308$). The 7B$\leftrightarrow$7B sub-population
from Run~2 is the correct scale-controlled comparison unit for the main
universality claim.}
\label{tab:b2_run_comparison}
\small
\begin{tabular}{lrrr}
\toprule
Metric & Run~1 & Run~2 & Run~2 7B$\leftrightarrow$7B \\
\midrule
Pass rate (\%)           & 68.4 & 45.5 & 48.9 \\
Proc.\ cos (mean)        & 0.592 & 0.641 & 0.753 \\
Proc.\ cos (median)      & 0.710 & 0.827 & 0.773 \\
$\rho_c$ fold enrichment (median) & 40.9$\times$ & 28.6$\times$ & --- \\
Cohen $d$ (mean)          & 1.213 & 0.423 & 0.442 \\
Pearson $r$ (mean)        & 0.651 & 0.512 & --- \\
SAE-free cos (mean)       & 0.893 & 0.925 & 0.927 \\
Fwd/rev symmetry          & N/A & $\leq$1.4pp & $\leq$1.4pp \\
\bottomrule
\end{tabular}
\end{table}

\begin{figure}[h!]
\centering
\includegraphics[width=\linewidth]{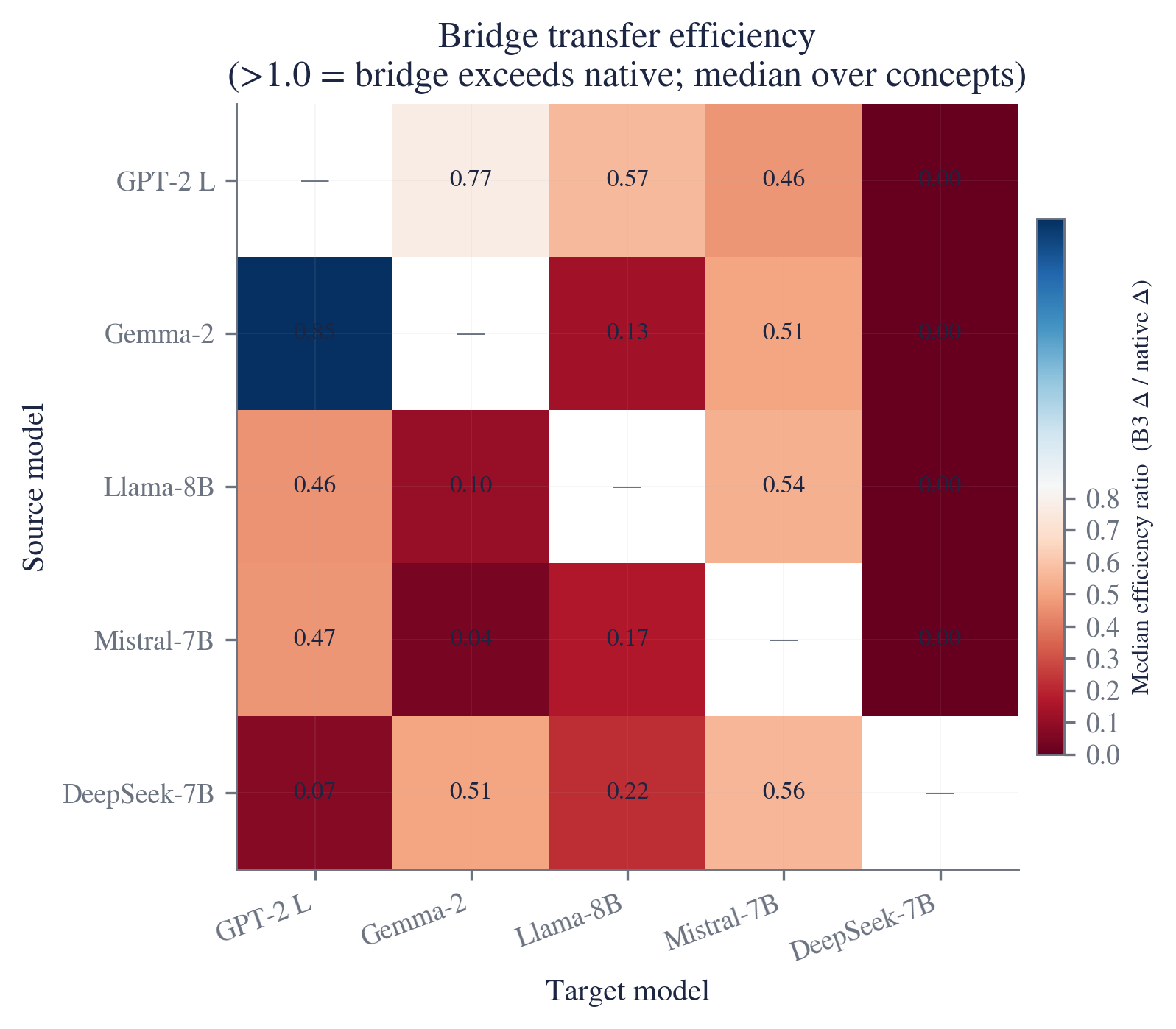}
\caption{Transfer efficiency heatmap: fraction of B2-validated feature pairs achieving each quality tier per directed model pair.}
\label{fig:transfer_eff}
\end{figure}

\begin{figure}[h!]
\centering
\includegraphics[width=\linewidth]{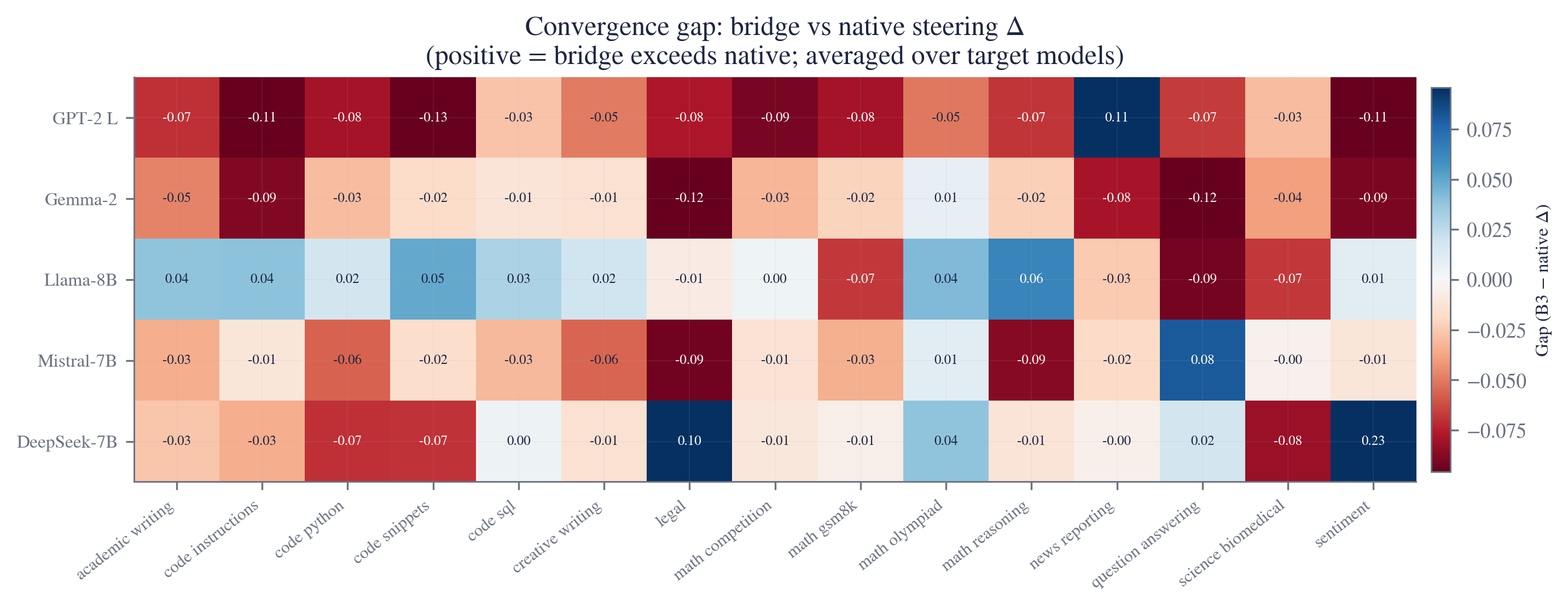}
\caption{Convergence quality gap: difference in pass rate between scale-matched (7B$\leftrightarrow$7B) and cross-scale (0.8B$\leftrightarrow$7B) model pairs per quality metric.}
\label{fig:convergence_gap}
\end{figure}

\begin{figure}[h!]
\centering
\includegraphics[width=\linewidth]{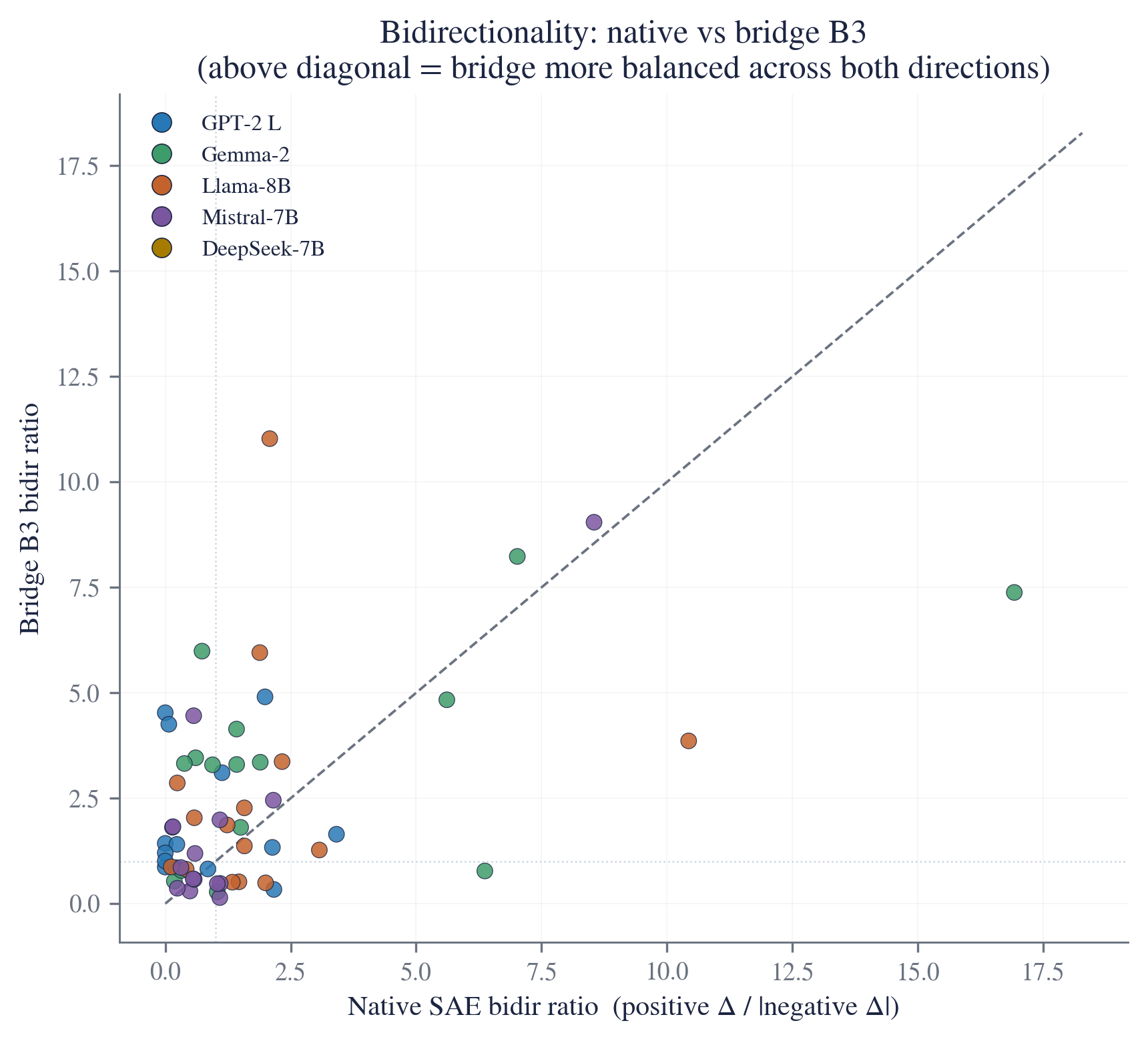}
\caption{Bidirectionality scatter: forward pass rate vs reverse pass rate for all 20 directed model pairs. Points near the diagonal confirm symmetry ($\leq$1.4pp gap for all pairs).}
\label{fig:bidir}
\end{figure}

\FloatBarrier
\section{Global MLP Training Details (C1)}
\label{app:c1}

The Global MLP (C1) has five per-model encoders
($d_{\text{model}} \to 2048 \to 512$, LayerNorm after each Linear, GELU,
Dropout~0.1) projecting into a shared 512-dimensional concept space, and
symmetric decoders ($512 \to 2048 \to d_{\text{model}}$) reconstructing the
original SAE activation vectors. Training runs for 200 epochs on
8$\times$A100-80GB with DistributedDataParallel (66 minutes); the loss
combines reconstruction MSE (weight 1.0) and NT-Xent contrastive loss
(weight 0.5, $\tau = 0.1$).

Table~\ref{tab:c1_epoch} reports the final training objective values;
Table~\ref{tab:c1_permodel} shows per-model validation reconstruction loss.

\begin{table}[h]
\centering
\caption{C1 Global MLP final training metrics (epoch 200).}
\label{tab:c1_epoch}
\small
\begin{tabular}{lrrrrrr}
\toprule
 & train total & train recon & train align & val total & val recon & val align \\
\midrule
Epoch 200 & 0.0665 & 0.0222 & 0.0887 & 0.0854 & 0.0133 & 0.1443 \\
\bottomrule
\end{tabular}
\end{table}

\begin{table}[h]
\centering
\caption{C1 per-model validation reconstruction loss (epoch 200).
gpt2-large achieves the lowest loss owing to its smaller SAE feature count
(666 features vs 815--883 for 7B models).}
\label{tab:c1_permodel}
\small
\begin{tabular}{lccccc}
\toprule
Model & gpt2-large & gemma-2-2b & llama-3.1-8B & mistral-7B & deepseek-7B \\
\midrule
Val recon & 0.00202 & 0.00540 & 0.01999 & 0.02249 & 0.01654 \\
\bottomrule
\end{tabular}
\end{table}

\noindent Zero concept-space neurons are dead (0.0\%) at epoch 200, confirming
all 512 shared dimensions actively encode information throughout training.
The validation alignment loss exceeds training alignment loss throughout
training (0.144 vs 0.089 at epoch 200), which is expected: contrastive
negative-pair sampling is harder at evaluation (no data augmentation), making
the in-batch contrast more difficult.

\paragraph{Instruction-tuning robustness note.}
The \dd{llama} model is \dd{Hermes-3-Llama-3.1-8B}, an instruction-tuned
variant, while the other four models are base checkpoints.
The per-model val recon for llama (0.020) is comparable to the base-model
7B entries (mistral 0.022, deepseek 0.017), indicating the Global MLP encodes
the instruction-tuned model's representations as faithfully as base models.
This is consistent with the B2 finding that llama's pairwise pass rates
(42.6--47.7\%) are within 2pp of same-scale base-model pairs,
suggesting supervised fine-tuning does not disrupt the underlying
universal concept geometry.

\section{Universal Concept Details (C2)}
\label{app:c2}

\subsection*{C2 Pipeline Overview}

This section reports the complete C2 universal concept discovery pipeline:
4,100 labelled SAE features (across 5 models) are projected through the C1
encoder into the 512-d shared concept space, reduced to 30 dimensions via
UMAP, clustered with HDBSCAN to produce 133 raw clusters, filtered to 113
clusters spanning $\geq$2 models, then canonically deduplicated to 11
universal concepts each with 5/5 model coverage.
Tables~\ref{tab:c2_input}--\ref{tab:c2_raw_labels} provide the full audit
trail. Table~\ref{tab:concepts} is the final canonical summary.

\subsection*{C2.1 Feature Input Per Model}

Table~\ref{tab:c2_input} reports the number of labelled SAE features per model
that serve as input to the C2 pipeline. These are the features selected during
A4a (top 150 per domain by activation delta, plus top 100 by mean activation
regardless of domain), labelled by Claude, and projected through the C1 encoder.
The unsupervised A4b auto-discovery was applied to gpt2-large only (48 additional
clusters from 81,920 SAE features on 10,000 passages); those 816 auto-discovered
feature entries are merged with the 666 supervised entries for gpt2-large in the
label file but only the 666 A4a entries are carried into C2 (auto-discovered
features lack semantic domain anchors needed for concept-space alignment).

\begin{table}[h]
\centering
\caption{Labelled SAE feature input per model to the C2 discovery pipeline
(A4a supervised labels only). All entries are projected through the
trained C1 encoder to their 512-d concept-space coordinates before UMAP
and HDBSCAN.}
\label{tab:c2_input}
\small
\begin{tabular}{lrrrr}
\toprule
Model & SAE $n_\text{features}$ & Expansion & Labelled (A4a) & In C2 \\
\midrule
gpt2-large   &  81,920 &  64$\times$ &  666 &  666 \\
gemma-2-2b   & 147,456 &  64$\times$ &  865 &  865 \\
llama-8B     & 524,288 & 128$\times$ &  871 &  871 \\
mistral-7B   & 524,288 & 128$\times$ &  883 &  883 \\
deepseek-7B  & 524,288 & 128$\times$ &  815 &  815 \\
\midrule
\textbf{Total} & --- & --- & \textbf{4,100} & \textbf{4,100} \\
\bottomrule
\end{tabular}
\end{table}

\subsection*{C2.2 HDBSCAN Run History}

Table~\ref{tab:c2_runs} summarises the three C2 runs. Runs~1--2 applied
HDBSCAN directly to the 512-dimensional shared space; the curse of
dimensionality causes pairwise distances to concentrate near their mean,
eliminating the density gradients HDBSCAN requires and producing 58--65\%
noise~\citep{beyer1999nearest}. Run~3 prepends UMAP reduction to 30 dimensions
($n_{\text{neighbors}}=50$, cosine metric, \texttt{min\_dist}=0), reducing
noise to 15.8\% and recovering 133 clusters.
The UMAP collapse to 30d preserves local neighbourhood structure
(McInnes et al.~\citeyear{mcinnes2018umap}); the \texttt{min\_dist}=0 setting
tightens clusters for HDBSCAN downstream, which is standard practice.
All three runs use \texttt{min\_cluster\_size}=20 and \texttt{min\_models}=2.

\begin{table}[h]
\centering
\caption{C2 HDBSCAN run history. UMAP pre-processing (512$\to$30d) resolves
the curse of dimensionality and makes clustering tractable. All runs use
\texttt{min\_cluster\_size}=20.}
\label{tab:c2_runs}
\small
\begin{tabular}{lcccccc}
\toprule
Run & UMAP & Clusters & Noise (\%) & Universal ($\geq$2 models) & After string dedup & Canonical \\
\midrule
1 & None &  8 & 58.0 &  3 & --- & --- \\
2 & None &  8 & 65.0 &  3 & --- & --- \\
3 & 512$\to$30d & 133 & 15.8 & 113 & 18 & \textbf{11} \\
\bottomrule
\end{tabular}
\end{table}

\subsection*{C2.3 Sub-Cluster Model Coverage}

Table~\ref{tab:c2_coverage} reports model coverage of the 107 non-noise raw
clusters (133 total $-$ 26 noise clusters) from Run~3 before semantic
deduplication. A cluster's coverage is the number of distinct models
that contribute at least one SAE feature to that cluster.

\begin{table}[h]
\centering
\caption{Model coverage distribution of the 107 non-noise raw HDBSCAN
clusters (Run~3) before semantic deduplication. The 5/5 union coverage
of all 11 canonical concepts is achieved through merging --- no individual
canonical concept requires every one of its contributing raw clusters to
be 5/5.}
\label{tab:c2_coverage}
\small
\begin{tabular}{crrl}
\toprule
Coverage & Clusters & \% & Interpretation \\
\midrule
5/5 & 46 & 43.0 & Core concept --- present in every architecture \\
4/5 & 41 & 38.3 & Strong concept --- one architecture diverges slightly \\
3/5 & 18 & 16.8 & Peripheral sub-cluster --- still cross-architecture \\
2/5 &  2 &  1.9 & Minimum threshold (clusters 24, 59) \\
\midrule
$\geq$4/5 & 87 & 81.3 & Robust pre-merge coverage \\
\bottomrule
\end{tabular}
\end{table}

The 2/5 minimum-threshold clusters (cluster 24: \texttt{math\_problem\_solving},
cluster 59: \texttt{legal\_criminal\_proceedings}) contribute to the 5/5 union
of their respective canonical concepts (\texttt{math\_problems} and
\texttt{legal\_and\_news}) after merging. Reviewers citing the 2/5 minimum
should note that the union-level coverage reported in Table~\ref{tab:concepts}
is the scientifically correct unit: a canonical concept is universal if every
model architecture contributes at least one sub-cluster, not if every
sub-cluster spans all five models.

\subsection*{C2.4 Universal $\to$ Deduplication: 113 Raw Clusters to 11 Concepts}

Of the 113 universal raw clusters, string-exact deduplication yields 18 unique
labels. This inflates the concept count because Claude generates syntactically
varied labels for semantically identical content
(\texttt{math\_word\_problems}, \texttt{mathematical\_problem\_solving},
\texttt{mathematical\_word\_problems}, \texttt{math\_problem\_solving},
\texttt{math\_problem\_solutions} all refer to the same concept).
A human canonical mapping step (performed by the authors) merges clusters
sharing the same underlying content, retaining the most frequent label as the
canonical name. Six multi-domain clusters without a coherent semantic focus are
marked as noise and excluded from C3. The final result is 11 canonical concepts,
each with 5/5 union coverage.

Table~\ref{tab:c2_dedup} shows the canonical mapping: each row is a final
concept, with the number of raw HDBSCAN clusters merged into it and the total
SAE feature member count.

\begin{table}[h]
\centering
\caption{Reduction from 113 universal raw HDBSCAN clusters to 11 canonical
concepts. Raw clusters: number of Run~3 clusters merged into this concept.
Features: total SAE feature members across all merged clusters (summed over
all 5 models). Representative raw labels are a sample; full label list is
in Table~\ref{tab:c2_raw_labels}.}
\label{tab:c2_dedup}
\resizebox{\columnwidth}{!}{%
\small
\begin{tabular}{lrcrl}
\toprule
Canonical concept & Raw clusters & Features & Models & Sample of raw Claude labels \\
\midrule
\texttt{python\_code}           & 29 & 185 & 5/5 & \texttt{python\_code\_headers}, \texttt{code\_with\_comments}, \texttt{python\_file\_headers} \\
\texttt{math\_problems}         & 20 & 389 & 5/5 & \texttt{math\_word\_problems}, \texttt{mathematical\_problem\_solving}, \texttt{math\_problem\_solutions} \\
\texttt{sql\_queries}           & 11 & 312 & 5/5 & \texttt{sql\_query\_examples}, \texttt{code\_and\_sql\_queries}, \texttt{sql\_query\_structure} \\
\texttt{legal\_and\_news}        & 10 & 267 & 5/5 & \texttt{legal\_policy\_news}, \texttt{legal\_news\_articles}, \texttt{legal\_government\_proceedings} \\
\texttt{medical\_research}       &  8 & 298 & 5/5 & \texttt{medical\_research\_conclusions}, \texttt{medical\_research\_findings}, \texttt{research\_study\_conclusions} \\
\texttt{academic\_scientific}    &  8 & 487 & 5/5 & \texttt{academic\_technical\_prose}, \texttt{academic\_research\_abstracts}, \texttt{scientific\_news\_abstracts} \\
\texttt{narrative\_fiction}      &  6 & 156 & 5/5 & \texttt{first\_person\_narrative}, \texttt{narrative\_storytelling\_prose}, \texttt{dialogue\_and\_narrative\_prose} \\
\texttt{encyclopedic\_historical}&  5 & 178 & 5/5 & \texttt{historical\_biographical\_events}, \texttt{encyclopedic\_factual\_statements}, \texttt{past\_events\_narrative} \\
\texttt{code\_and\_math}$^\star$        &  4 & 401 & 5/5 & \texttt{code\_and\_math}, \texttt{math\_and\_code}, \texttt{mathematical\_code\_problems} \\
\texttt{customer\_reviews}       &  3 & 224 & 5/5 & \texttt{restaurant\_reviews}, \texttt{customer\_reviews\_feedback}, \texttt{personal\_experience\_reviews} \\
\texttt{sql\_and\_medical}$^\star$      &  3 & 203 & 5/5 & \texttt{sql\_and\_medical}, \texttt{medical\_research\_and\_sql}, \texttt{technical\_query\_statements} \\
\midrule
\textbf{Total (11 concepts)} & \textbf{107} & \textbf{3,100} & \textbf{5/5} & \\
\midrule
NOISE (excluded)             &   6 & --- & --- & \texttt{multi\_domain\_text\_samples}, \texttt{incomplete\_text\_continuations}, \texttt{technical\_formal\_prose} \\
\bottomrule
\end{tabular}
}
\end{table}

$^\star$ Compound concepts (\texttt{code\_and\_math}, \texttt{sql\_and\_medical}) emerge from
cross-domain co-activation in the shared space and are not present in the
original 15 supervised domains. They suggest models represent higher-order
compositional structure --- joint text-type selectivity --- not only
domain-level features.

\subsection*{C2.5 Raw Claude Label Distribution (40 Distinct Labels)}

Table~\ref{tab:c2_raw_labels} is the complete audit trail of all 40 distinct
Claude-generated label strings from the 113 universal clusters. It documents
which labels were merged, justifying the reduction from 40 strings to 11
canonical concepts and demonstrating that string-exact deduplication alone
cannot recover the true concept count.

\begin{table}[h]
\centering
\caption{All 40 raw Claude label strings from the 113 universal HDBSCAN clusters
(Run~3, temperature=0). Labels are deterministic: same passages always produce
the same label. ``Clusters'' = number of HDBSCAN clusters producing this label.
``Coverage'' = modal model coverage over those clusters.
$\to$ Canonical = final canonical concept after human semantic deduplication.}
\label{tab:c2_raw_labels}
\small
\begin{tabular}{lccl}
\toprule
Raw Claude label & Clusters & Coverage & $\to$ Canonical concept \\
\midrule
\texttt{python\_code\_headers}           & 12 & 4--5/5 & \texttt{python\_code} \\
\texttt{math\_word\_problems}            & 12 & 3--5/5 & \texttt{math\_problems} \\
\texttt{medical\_research\_conclusions}  &  5 & 3--5/5 & \texttt{medical\_research} \\
\texttt{code\_and\_sql\_queries}         &  4 & 4--5/5 & \texttt{sql\_queries} \\
\texttt{sql\_query\_examples}            &  3 & 4--5/5 & \texttt{sql\_queries} \\
\texttt{mathematical\_problem\_solving}  &  3 & 3--5/5 & \texttt{math\_problems} \\
\texttt{first\_person\_narrative}        &  3 & 3--5/5 & \texttt{narrative\_fiction} \\
\texttt{legal\_policy\_news}             &  3 & 4/5    & \texttt{legal\_and\_news} \\
\texttt{code\_and\_sql}                  &  2 & 5/5    & \texttt{sql\_queries} \\
\texttt{code\_and\_technical\_text}      &  2 & 5/5    & \texttt{python\_code} \\
\texttt{academic\_technical\_prose}      &  2 & 4/5    & \texttt{academic\_scientific} \\
\texttt{code\_with\_comments}            &  2 & 5/5    & \texttt{python\_code} \\
\texttt{python\_file\_headers}           &  2 & 5/5    & \texttt{python\_code} \\
\texttt{legal\_news\_articles}           &  2 & 3--5/5 & \texttt{legal\_and\_news} \\
\texttt{legal\_and\_mathematical\_problems} & 2 & 3/5  & NOISE \\
\texttt{truncated\_text\_passages}       &  2 & 3/5    & NOISE \\
\texttt{math\_problem\_solving}          &  2 & 2--3/5 & \texttt{math\_problems} \\
\texttt{python\_code\_and\_documentation} & 1 & 4/5   & \texttt{python\_code} \\
\texttt{python\_script\_headers}         &  1 & 5/5    & \texttt{python\_code} \\
\texttt{python\_import\_statements}      &  1 & 3/5    & \texttt{python\_code} \\
\texttt{python\_imports\_headers}        &  1 & 4/5    & \texttt{python\_code} \\
\texttt{python\_imports\_and\_legal}     &  1 & 3/5    & \texttt{python\_code} \\
\texttt{software\_license\_headers}      &  1 & 4/5    & \texttt{python\_code} \\
\texttt{code\_comments\_and\_headers}    &  1 & 5/5    & \texttt{python\_code} \\
\texttt{code\_and\_technical}            &  1 & 4/5    & \texttt{python\_code} \\
\texttt{code\_and\_technical\_data}      &  1 & 4/5    & \texttt{python\_code} \\
\texttt{code\_and\_legal\_text}          &  1 & 3/5    & \texttt{python\_code} \\
\texttt{code\_and\_math}                 &  1 & 5/5    & \texttt{code\_and\_math} \\
\texttt{math\_and\_code}                 &  1 & 5/5    & \texttt{code\_and\_math} \\
\texttt{code\_and\_math\_problems}       &  1 & 3/5    & \texttt{code\_and\_math} \\
\texttt{mathematical\_code\_problems}    &  1 & 5/5    & \texttt{code\_and\_math} \\
\texttt{math\_problem\_solutions}        &  1 & 3/5    & \texttt{math\_problems} \\
\texttt{mathematical\_word\_problems}    &  1 & 3/5    & \texttt{math\_problems} \\
\texttt{mathematical\_and\_administrative\_text} & 1 & 5/5 & \texttt{math\_problems} \\
\texttt{sql\_query\_structure}           &  1 & 3/5    & \texttt{sql\_queries} \\
\texttt{sql\_query\_comments}            &  1 & 5/5    & \texttt{sql\_queries} \\
\texttt{sql\_and\_code}                  &  1 & 4/5    & \texttt{sql\_queries} \\
\texttt{code\_and\_medical\_abstracts}   &  1 & 5/5    & NOISE \\
\texttt{technical\_formal\_prose}        &  1 & 5/5    & NOISE \\
\texttt{multi\_domain\_text\_samples}    &  1 & 5/5    & NOISE \\
\bottomrule
\end{tabular}
\end{table}

\noindent The high fragmentation of the \texttt{python\_code} label (29 clusters, 12 distinct label strings)
reflects that HDBSCAN discovers fine-grained code-register sub-clusters
(headers, imports, comments, license text) that the human canonical mapping
consolidates. The compound concepts (\texttt{code\_and\_math},
\texttt{sql\_and\_medical}) have only 4 and 3 raw clusters respectively,
suggesting they are cohesive cross-domain regions rather than aggregated fragments.

\subsection*{C2.6 Canonical Universal Concepts (Final 11)}

Table~\ref{tab:concepts} reports the 11 canonical universal concepts with
model coverage, total SAE feature members, number of raw HDBSCAN clusters
merged, and mean C3 steering delta. All 11 have 5/5 model coverage (union
across member clusters).

\begin{table}[h]
\centering
\caption{The 11 canonical universal concepts from C2. Coverage: union of models
contributing at least one SAE feature. Features: total SAE features assigned to
this concept across all 5 models. Raw clusters: number of HDBSCAN Run-3
clusters merged. Mean C3 $\Delta$: mean signed concept-score delta at best
positive injection strength per (model, concept) pair. $^\star$ = compound
cross-domain concept absent from the original 15 supervised domains.}
\label{tab:concepts}
\small
\begin{tabular}{lcccr}
\toprule
Canonical concept & Coverage & Features & Raw clusters & Mean C3 $\Delta$ \\
\midrule
\texttt{academic\_scientific}     & 5/5 & 487 &  8 & $+0.142$ \\
\texttt{sql\_queries}             & 5/5 & 312 & 11 & $+0.132$ \\
\texttt{code\_and\_math}$^\star$  & 5/5 & 401 &  4 & $+0.113$ \\
\texttt{medical\_research}        & 5/5 & 298 &  8 & $+0.113$ \\
\texttt{legal\_and\_news}         & 5/5 & 267 & 10 & $+0.111$ \\
\texttt{math\_problems}           & 5/5 & 389 & 20 & $+0.089$ \\
\texttt{sql\_and\_medical}$^\star$ & 5/5 & 203 &  3 & $+0.108$ \\
\texttt{encyclopedic\_historical} & 5/5 & 178 &  5 & $+0.068$ \\
\texttt{customer\_reviews}        & 5/5 & 224 &  3 & $+0.081$ \\
\texttt{narrative\_fiction}       & 5/5 & 156 &  6 & $+0.041$ \\
\texttt{python\_code}             & 5/5 & 185 & 29 & $+0.047$ \\
\midrule
\textbf{Total}                    & --- & \textbf{3,100} & \textbf{107} & $+0.095$ \\
\bottomrule
\end{tabular}
\end{table}

The three highest-performing concepts by mean C3~$\Delta$
(\texttt{academic\_scientific}, \texttt{sql\_queries}, \texttt{code\_and\_math})
are structural registers in which token-level distributional signals are
strong and consistent (specialised vocabulary, syntax, and formatting).
The two lowest-performing (\texttt{narrative\_fiction}, \texttt{python\_code})
have more diffuse lexical profiles --- narrative prose is maximally
context-dependent and code headers are dominated by repeated boilerplate
tokens that suppress concept-score variance relative to the DeBERTa baseline.

\subsection*{C2.7 Notable Observation: Cluster 28 (Structural Universality)}

One cluster excluded from the 11 canonical concepts merits specific mention.
Cluster~28 (label \texttt{incomplete\_text\_continuations}, 5/5 models,
excluded as noise because evidence passages were malformed) represents
\emph{syntactic incompleteness} --- all five models independently developed a
dedicated representation for mid-sentence truncated text, independent of
content domain (math, code, prose). This is a structural rather than semantic
concept: the universal information encoded is ``text that stops abruptly.''
It is excluded from C3 steering vectors because there is no coherent semantic
direction to steer toward, but it is reported here because it demonstrates that
cross-architecture universality extends to structural syntactic properties, not
only to semantic content domains.

\begin{figure}[h!]
\centering
\begin{minipage}[t]{0.48\linewidth}
  \includegraphics[width=\linewidth]{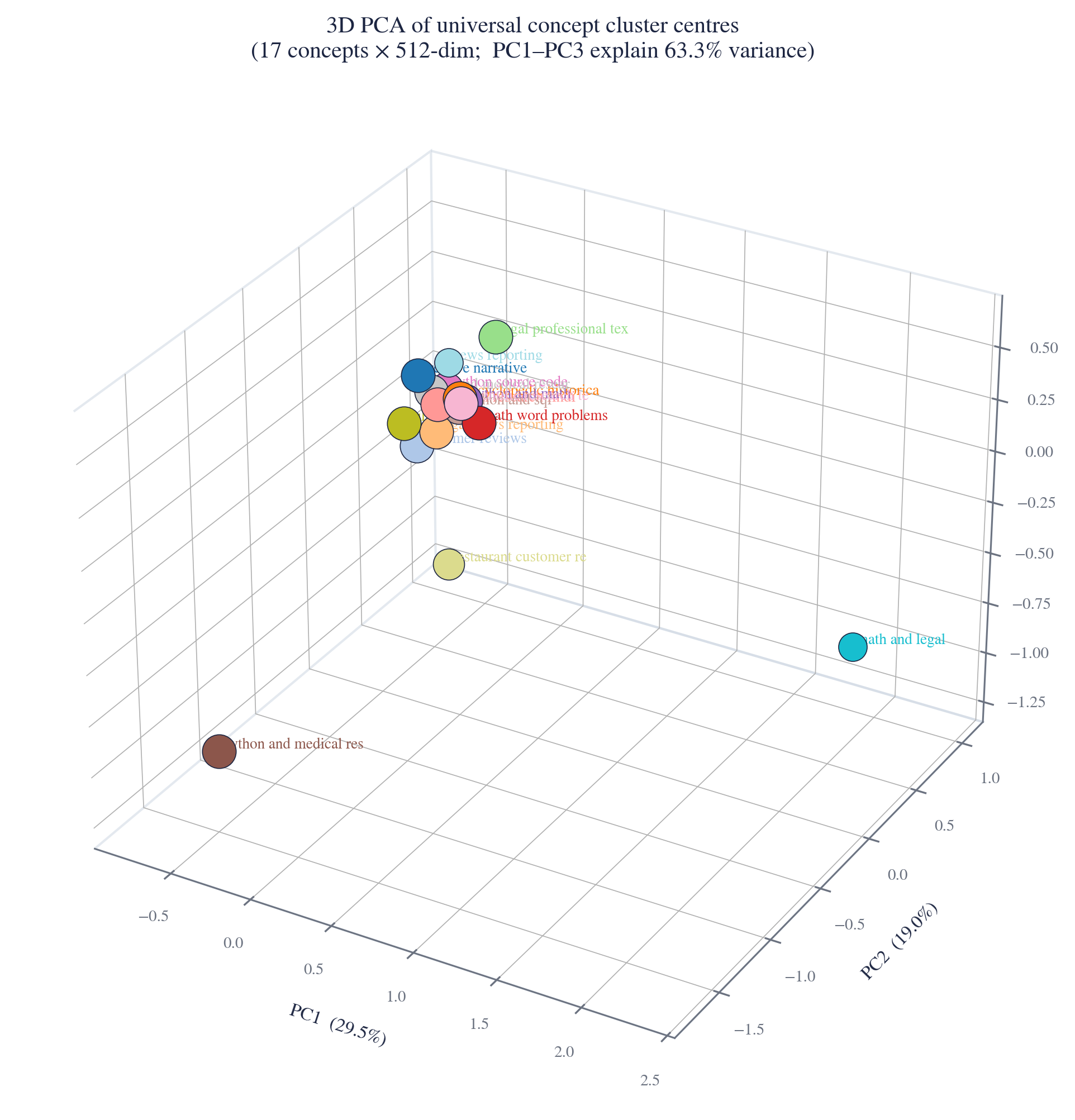}
  \subcaption{3D view (PC1--PC3, 63.3\% variance)}
  \label{fig:clusters_3d}
\end{minipage}\hfill
\begin{minipage}[t]{0.48\linewidth}
  \includegraphics[width=\linewidth]{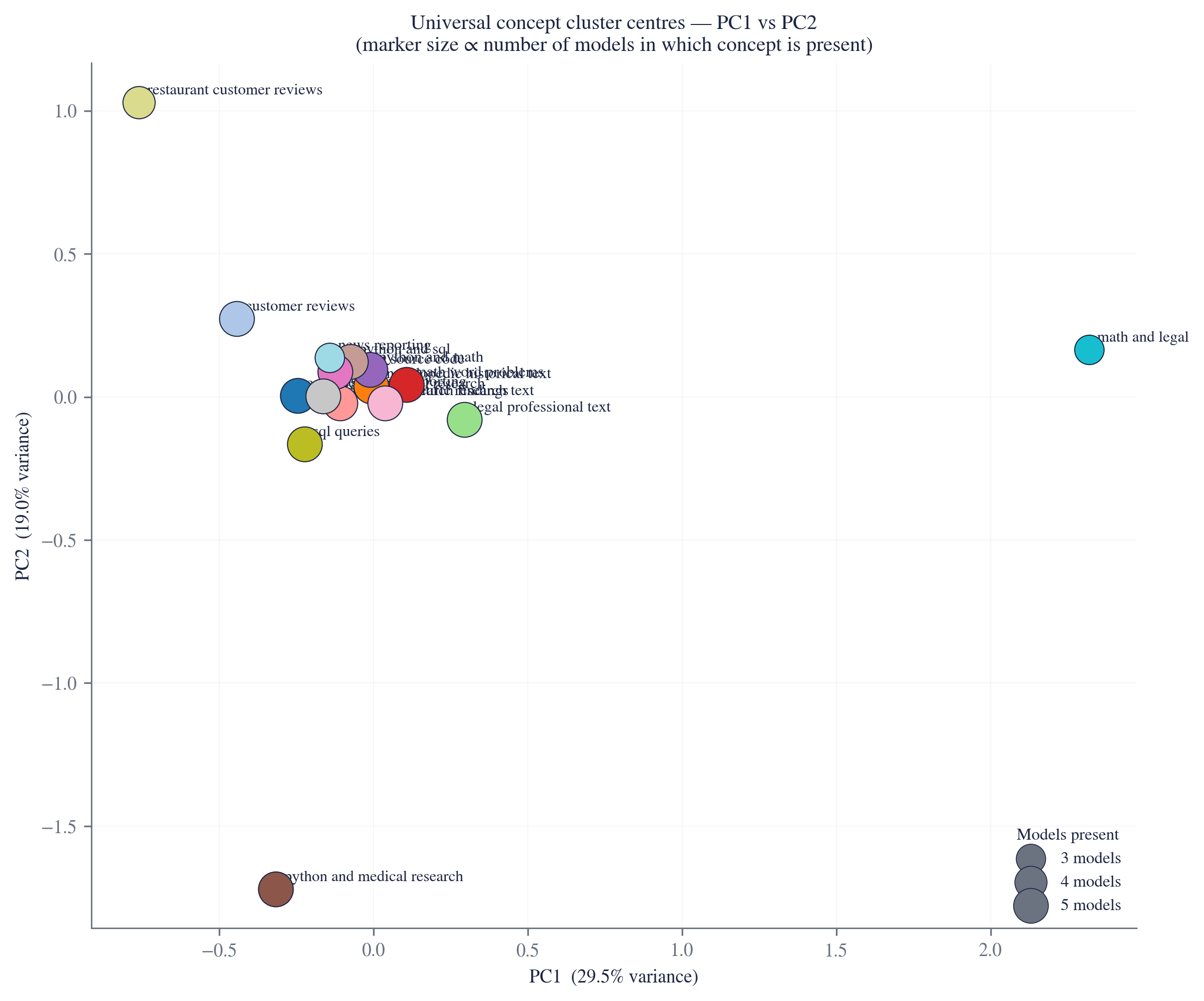}
  \subcaption{PC1 vs PC2; marker size $\propto$ models present}
  \label{fig:clusters_2d}
\end{minipage}
\caption{PCA of 17 universal concept cluster centres (512-dim). Three outlier concepts --- \textit{python\_and\_medical\_research}, \textit{restaurant\_customer\_reviews}, and \textit{math\_and\_legal} --- are geometrically distant from the main cluster, suggesting they span distinct representational axes. PC1--PC3 explain 63.3\% of variance.}
\label{fig:clusters_pca}
\end{figure}

\section{Hallucination and Stability Analysis}
\label{app:hallucination}

Table~\ref{tab:hallucination} reports repetition rates (fraction of outputs with
rate $>$ 0) by model, method, and strength.

\begin{table}[h]
\centering
\caption{Repetition rate (\%) and valid-output rate (\%) by model and
injection strength ($s$). Outputs with repetition rate $>0.40$ are excluded from
all reported means (rep gate). deepseek is evaluated at $s=1$ only in the main
analysis (valid-output rate $<$10\% at $s\geq 2$); gpt2 results are restricted
to $s\leq 3$ (valid rate falls to 33\% at $s=5$).}
\label{tab:hallucination}
\small
\begin{tabular}{llcccc}
\toprule
Model & Method & $s=1$ & $s=2$ & $s=3$ & $s=5$ \\
\midrule
deepseek  & sae\_vector & 53 & 84 & 88 & 97 \\
deepseek  & b3\_ti      & 81 & 100 & 100 & 100 \\
gemma     & sae\_vector & 13 & 19 & 23 & 29 \\
gemma     & b3\_ti      & 14 & 16 & 15 & 29 \\
gpt2      & sae\_vector & 33 & 46 & 47 & 63 \\
gpt2      & b3\_ti      & 34 & 42 & 61 & 72 \\
llama     & sae\_vector &  0 &  1 &  1 &  5 \\
llama     & b3\_ti      &  0 &  2 &  2 &  2 \\
mistral   & sae\_vector &  6 &  9 & 12 & 13 \\
mistral   & b3\_ti      &  4 &  9 & 12 & 21 \\
\bottomrule
\end{tabular}
\end{table}

\begin{table}[h]
\centering
\caption{Valid-output rate (\%) by model, method, and injection strength ($s$).
Valid = repetition rate $\leq 0.40$. Derived directly from the gate applied
to all reported means. Cells marked $\dagger$ are not used in the main analysis
due to valid-output rate below the 25\% usability threshold.}
\label{tab:valid_rate}
\small
\begin{tabular}{llcccc}
\toprule
Model & Method & $s=1$ & $s=2$ & $s=3$ & $s=5$ \\
\midrule
deepseek  & sae\_vector & 47 & 16$^\dagger$ & 12$^\dagger$ & ~~3$^\dagger$ \\
deepseek  & b3\_ti      & 19$^\dagger$ & ~~0$^\dagger$ & ~~0$^\dagger$ & ~~0$^\dagger$ \\
gemma     & sae\_vector & 87 & 81          & 77          & 71 \\
gemma     & b3\_ti      & 86 & 84          & 85          & 71 \\
gpt2      & sae\_vector & 67 & 54          & 53          & 37$^\dagger$ \\
gpt2      & b3\_ti      & 66 & 58          & 39$^\dagger$ & 28$^\dagger$ \\
llama     & sae\_vector & 100 & 99         & 99          & 95 \\
llama     & b3\_ti      & 100 & 98         & 98          & 98 \\
mistral   & sae\_vector & 94 & 91          & 88          & 87 \\
mistral   & b3\_ti      & 96 & 91          & 88          & 79 \\
\bottomrule
\end{tabular}
\end{table}

\noindent Valid-output rates confirm the per-model analysis scope: deepseek
drops below 25\% usability at $s \geq 2$ for sae\_vector and at all strengths
for b3\_ti, motivating its restriction to $s=1$ in the main analysis.
gpt2 falls below 25\% for b3\_ti at $s \geq 3$ and for sae\_vector at $s=5$,
motivating its restriction to $s \leq 3$. llama, gemma, and mistral maintain
$\geq 71\%$ valid-output rate across all evaluated strengths.

\begin{figure*}[h!]
\centering
\includegraphics[width=\textwidth]{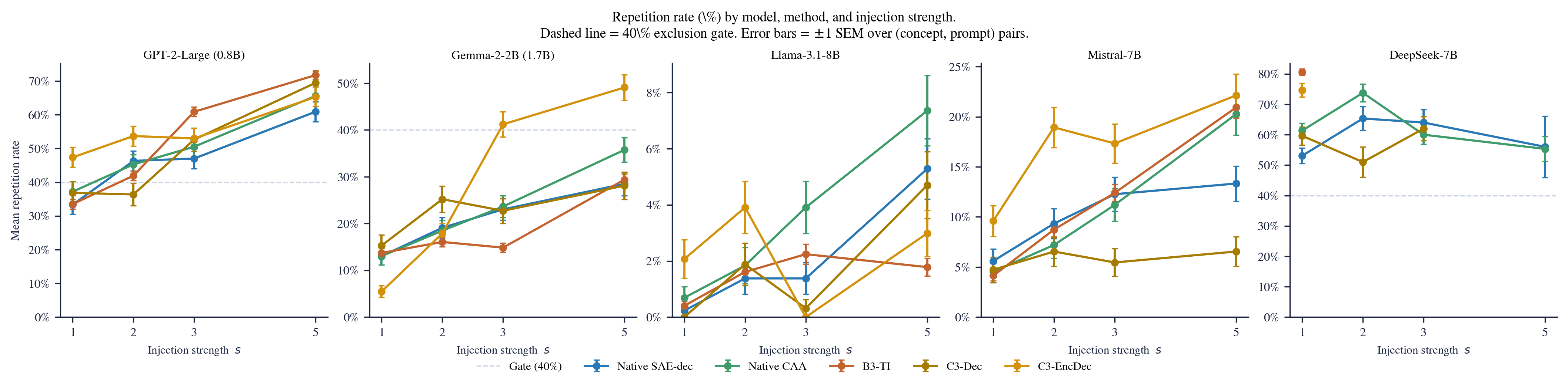}
\caption{Repetition rate (\%) as a function of injection strength $s$ per model
  and method (5 sub-panels). Error bars show $\pm$1~SEM over (concept, prompt)
  pairs. Dashed horizontal line marks the 40\% exclusion gate applied to all
  reported results. \dd{deepseek-llm-7b} exceeds the gate at $s\geq 2$ for
  all methods; \dd{gpt2-large} exceeds it for B3-TI at $s\geq 3$. \dd{llama},
  \dd{gemma}, and \dd{mistral} remain well below the gate at all tested
  strengths, confirming these three models as the stable evaluation set.}
\label{fig:hallucination_per_model}
\end{figure*}

\FloatBarrier
\section{Steering Vector Geometry}
\label{app:vec_geometry}

\subsection*{Concept Separability: C3 vs Enc-Dec}

\begin{figure*}[h!]
\centering
\includegraphics[width=\textwidth]{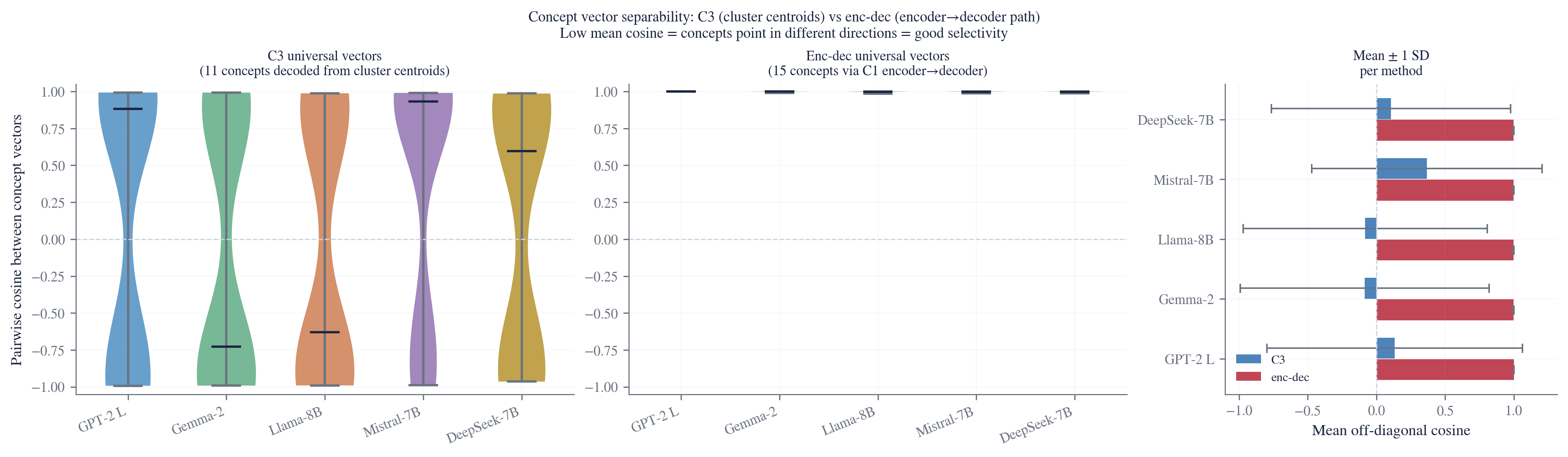}
\caption{Concept vector separability: pairwise off-diagonal cosine distributions
across all 11 C3 concepts (left) and all 15 enc-dec concepts (centre) per model,
plus per-method mean $\pm$ 1 SD (right). C3 vectors span the full unit sphere
(cosines from $-0.99$ to $+0.99$); enc-dec vectors collapse into a narrow cone
(cosines $0.985$--$1.000$) because the C1 encoder, trained on passage
co-activations, does not preserve concept-discriminative directional signal.}
\label{fig:concept_separability}
\end{figure*}

Figure~\ref{fig:concept_separability} visualises why C3 and enc-dec differ in
concept selectivity despite sharing the same Global MLP backbone. For C3,
concept vectors are decoded directly from the 11 HDBSCAN cluster centroids in
the shared 512-d space; these centroids are well-separated by construction
(cross-concept cosine range $[-0.99, +0.99]$, mean near zero across all five
models), so each decoded native vector points in a genuinely distinct direction.
For enc-dec, a guide model's native CAA steering direction is first\emph{encoded}
through the C1 encoder into the shared space, then decoded for the target.
Because the C1 encoder was trained with NT-Xent contrastive loss on
\emph{passage activations} (not on concept directions), it compresses all
incoming directions into the subspace it learned to represent passage-level
co-activations. The result is that all 15 enc-dec vectors cluster in a narrow
cone in the target model's activation space (cross-concept cosine
$0.985$--$0.9999$ depending on model), with near-zero variance across concepts.
The encoder path faithfully transports passage-level geometry but does not
preserve the concept-discriminative signal present in the original native
directions. This explains enc-dec's lower concept-score delta in Table~\ref{tab:transfer}
(65.3\% vs 67.3\% for C3): the decoded vectors are not well-separated enough
to exert concept-specific causal pressure.

\subsection*{C3 Concept Geometry per Model}

\begin{figure*}[h!]
\centering
\includegraphics[width=\textwidth]{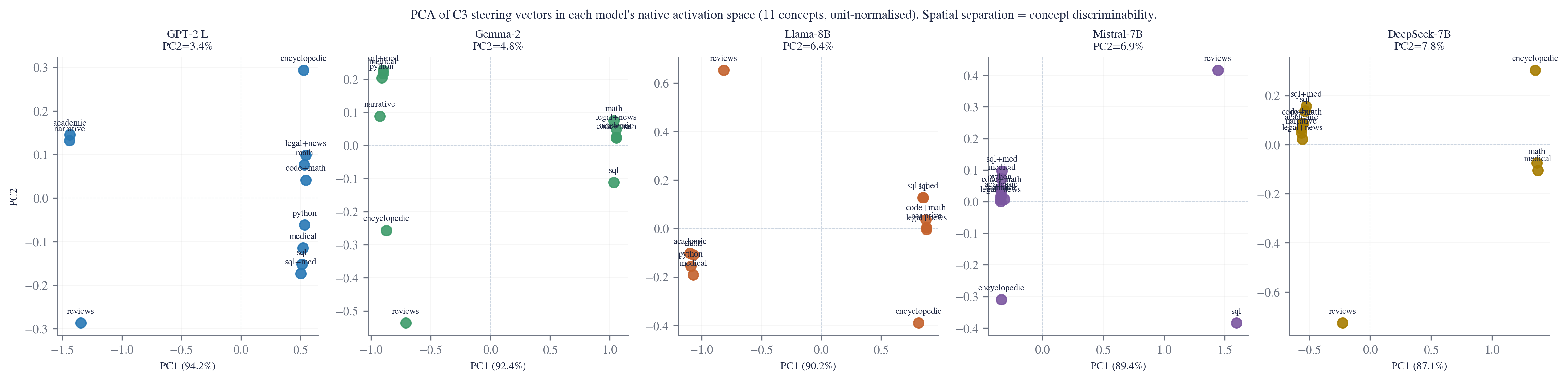}
\caption{PCA of 11 C3 steering vectors in each model's native activation space
(unit-normalised, 5 panels). Each point is one concept. Orthogonal spread
confirms well-separated concept directions; semantic groupings (code/math,
legal/news, medical) are visible in all five architectures.}
\label{fig:c3_pca_models}
\end{figure*}

Figure~\ref{fig:c3_pca_models} shows a 2D PCA of the 11 C3 steering vectors
in each model's native activation space. Concepts are well-separated across
all five architectures, with semantic clusters visible: code/math and SQL
concepts sit near each other, while medical and narrative fiction occupy
opposite poles. The per-model geometry varies in orientation but preserves
relative concept distances, consistent with the PCA of cluster centres
(Figure~\ref{fig:clusters_pca}) showing that the shared concept space has a
stable geometric structure that survives the per-model decoder step.

\subsection*{B3-TI Source Consistency}

\begin{figure*}[h!]
\centering
\includegraphics[width=\textwidth]{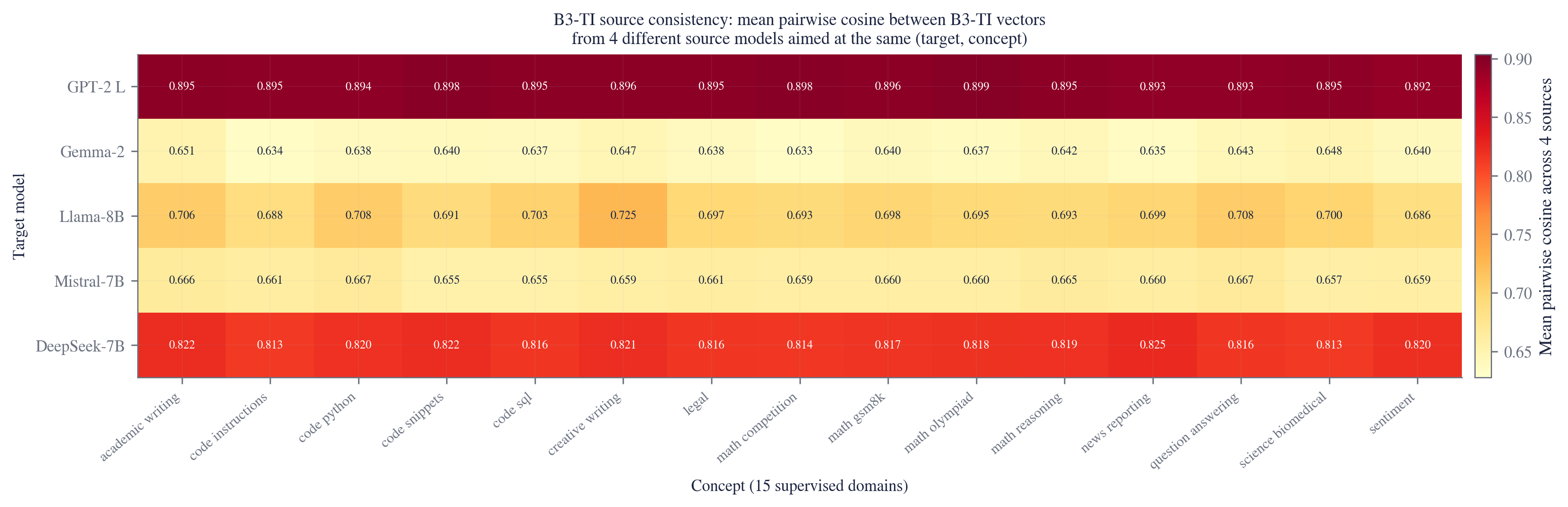}
\caption{B3-TI source consistency heatmap: mean pairwise cosine between the
four B3-TI vectors (from four different source models) targeted at each
(target model, concept) cell. High values indicate all source models
agree on which direction to push the target. \dd{gpt2-large} (0.8B) shows
the highest consistency (0.895), consistent with its smaller, more
constrained representational space.}
\label{fig:ti_source_consistency}
\end{figure*}

Figure~\ref{fig:ti_source_consistency} reports the mean pairwise cosine between
the four B3-TI vectors aimed at the same (target model, concept) from four
different source models. If the MLP bridge learns a faithful and consistent
translation, vectors from different sources should converge on the same
target direction. For \dd{gpt2-large} as target (mean 0.895), source
vectors agree strongly --- the 0.8B representational space imposes a
tight constraint that all four bridges converge towards. For 7B targets
(means 0.64--0.70), the larger target space allows more source-dependent
variation. The pattern mirrors the scale-tier hierarchy in
Table~\ref{tab:scale_tiers}: the same geometric bottleneck that limits
cross-scale alignment also makes the target space more \'reachable\' from
any source.

\FloatBarrier
\section{Prompt-Sensitivity Distribution}
\label{app:prompt_distrib}

\begin{figure*}[h!]
\centering
\includegraphics[width=\textwidth]{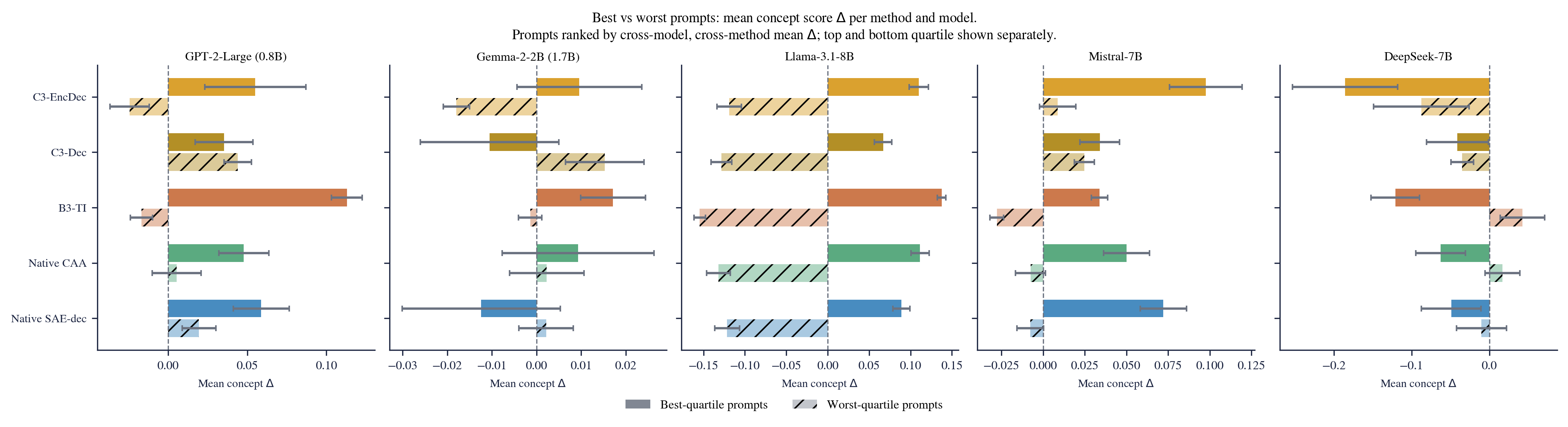}
\caption{Best vs worst prompts per model and method (5 sub-panels).
  Solid bars: mean concept-score $\Delta$ for best-quartile prompts
  (top 25\% ranked by cross-method mean $\Delta$);
  hatched bars: worst-quartile prompts.
  All five methods shown per model. The gap between best and worst quartiles
  is largest for \dd{mistral-7b} and \dd{gemma-2-2b} and smallest for
  \dd{deepseek-llm-7b} (restricted to $s=1$, compressing the dynamic range).
  This confirms that prompt register interacts strongly with steering
  efficacy, and that register-neutral prompts are necessary for fair
  cross-method comparison.}
\label{fig:prompt_best_worst_detail}
\end{figure*}

Of the 30 evaluation prompts, 14 meet the qualifying criterion (both B3-TI
and at least one universal method outperform same-model native vectors on the
averaged (model, concept) pair). Table~\ref{tab:prompt_list} lists these 14
prompts in descending order of mean concept-score delta at best positive strength.
The 4 prompts marked $\star$ are the only ones satisfying the stricter
``full hierarchy'' criterion (universal $>$ B3-TI $>$ native).
Figure~\ref{fig:prompt_best_worst_detail} shows that steerability is bimodal:
best-quartile prompts yield consistently positive $\Delta$ across all methods,
while worst-quartile prompts produce near-zero or negative deltas.

\begin{table}[h]
\centering
\caption{Qualifying prompts (14/30). Score = mean $\Delta$ across (model, concept) pairs at best positive strength. $\star$ = full hierarchy holds.}
\label{tab:prompt_list}
\small
\begin{tabular}{clc}
\toprule
Rank & Prompt stub & Score \\
\midrule
$\star$ 1 & Describe it & $+0.086$ \\
2 & The performance seems & $+0.072$ \\
3 & While both options have merit, & $+0.065$ \\
$\star$ 4 & Experts in the field agree that & $+0.050$ \\
5 & What's on your mind today & $+0.050$ \\
$\star$ 6 & Show me how & $+0.024$ \\
7 & Give me an example & $+0.024$ \\
8 & What do you think & $+0.023$ \\
9 & Explain how & $+0.022$ \\
10 & Researchers have consistently found that & $+0.021$ \\
11 & The consensus among researchers is that & $+0.019$ \\
$\star$ 12 & Broadly speaking, & $+0.017$ \\
13 & The study concluded that & $+0.014$ \\
14 & There are multiple ways to think about this & $+0.010$ \\
\bottomrule
\end{tabular}
\end{table}

Table~\ref{tab:supervised_full} reports mean signed delta, win rate, and
delta at $s=1$ for all evaluated methods on the full 30-prompt supervised
evaluation set.

\begin{table}[h]
\centering
\caption{Full supervised results (15 concepts, 30 prompts, 5 models).
(1) Mean signed $\Delta$ including negatives; (2) win rate (\% pairs with $\Delta>0$);
(3) $\Delta$ at fixed strength $s=1$.}
\label{tab:supervised_full}
\small
\begin{tabular}{lrrrrrr}
\toprule
Method & deepseek & gemma & gpt2 & llama & mistral & Mean \\
\midrule
\multicolumn{7}{l}{\emph{(1) Mean signed delta}} \\
Nat. SAE-dec & $+.013$ & $+.017$ & $+.039$ & $+.040$ & $+.040$ & $+.030$ \\
Native CAA  & $+.035$ & $+.015$ & $+.023$ & $+.036$ & $+.044$ & $+.031$ \\
B3-TI       & $-.012$ & $+.022$ & $+.037$ & $+.024$ & $+.031$ & $+.020$ \\
Naive       & $+.010$ & $+.023$ & $+.024$ & $+.014$ & $+.045$ & $+.023$ \\
C3-EncDec   & $-.001$ & $-.009$ & $+.030$ & $+.025$ & $+.072$ & $+.023$ \\
\midrule
\multicolumn{7}{l}{\emph{(2) Win rate (\%)}} \\
Nat. SAE-dec & 13.3 & 80.0 & 80.0 & 73.3 & 93.3 & 68.0 \\
Native CAA  & 26.7 & 73.3 & 53.3 & 86.7 & 86.7 & 65.3 \\
B3-TI       & 21.7 & 78.3 & 83.3 & 76.7 & 95.0 & 71.0 \\
Naive       & 16.7 & 80.0 & 70.0 & 71.7 & 90.0 & 65.7 \\
C3-EncDec   & ~~0.0 & 66.7 & 86.7 & 73.3 & 100.0 & 65.3 \\
\midrule
\multicolumn{7}{l}{\emph{(3) Delta at $s=1$}} \\
Nat. SAE-dec & $-.017$ & $-.005$ & $+.024$ & $-.011$ & $+.010$ & $.000$ \\
Native CAA  & $-.014$ & $+.004$ & $+.008$ & $-.010$ & $+.007$ & $-.001$ \\
B3-TI       & $-.033$ & $+.003$ & $+.012$ & $-.007$ & $+.006$ & $-.004$ \\
Naive       & $-.001$ & $+.001$ & $+.007$ & $-.009$ & $+.010$ & $+.002$ \\
C3-EncDec   & $-.001$ & $+.016$ & $-.035$ & $-.008$ & $+.010$ & $-.004$ \\
\bottomrule
\end{tabular}
\end{table}

\begin{figure*}[h!]
\centering
\includegraphics[width=\textwidth]{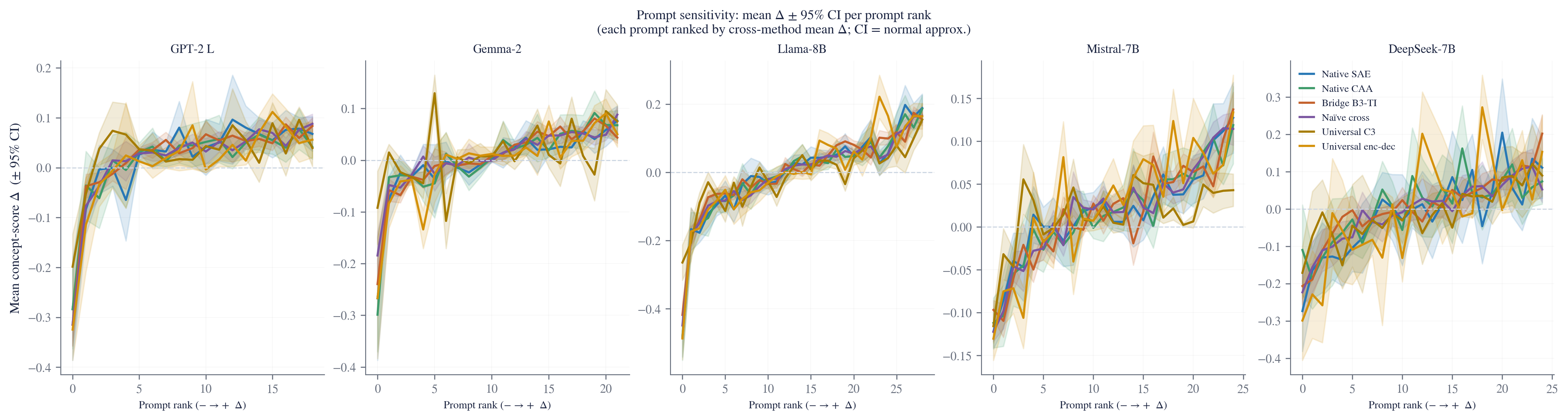}
\caption{Prompt sensitivity curves: mean concept-score~$\Delta \pm 95\%$~CI per prompt rank for each model and method class. Prompts ranked ascending by cross-method mean~$\Delta$. CI computed as normal approximation ($\pm 1.96 \cdot \mathrm{SE}$).}
\label{fig:prompt_sensitivity}
\end{figure*}

\begin{figure*}[h!]
\centering
\includegraphics[width=\textwidth]{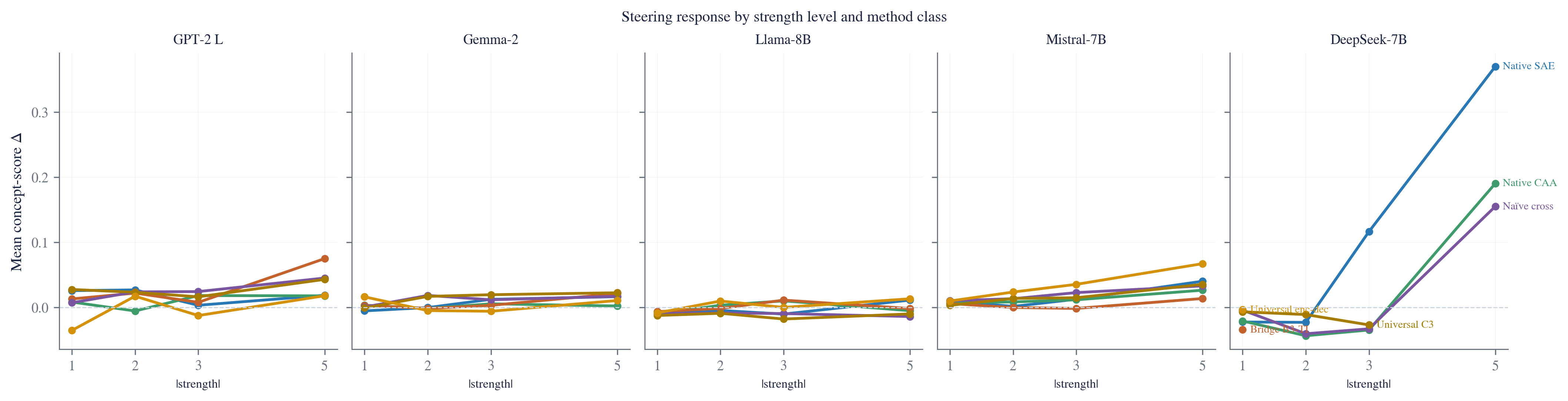}
\caption{Strength response curves: mean concept-score~$\Delta$ as a function of injection strength $s$ per model and method class. Lines show mean $\pm$ SE across (concept, prompt) pairs.}
\label{fig:strength_curves}
\end{figure*}

\begin{figure*}[h!]
\centering
\includegraphics[width=\textwidth]{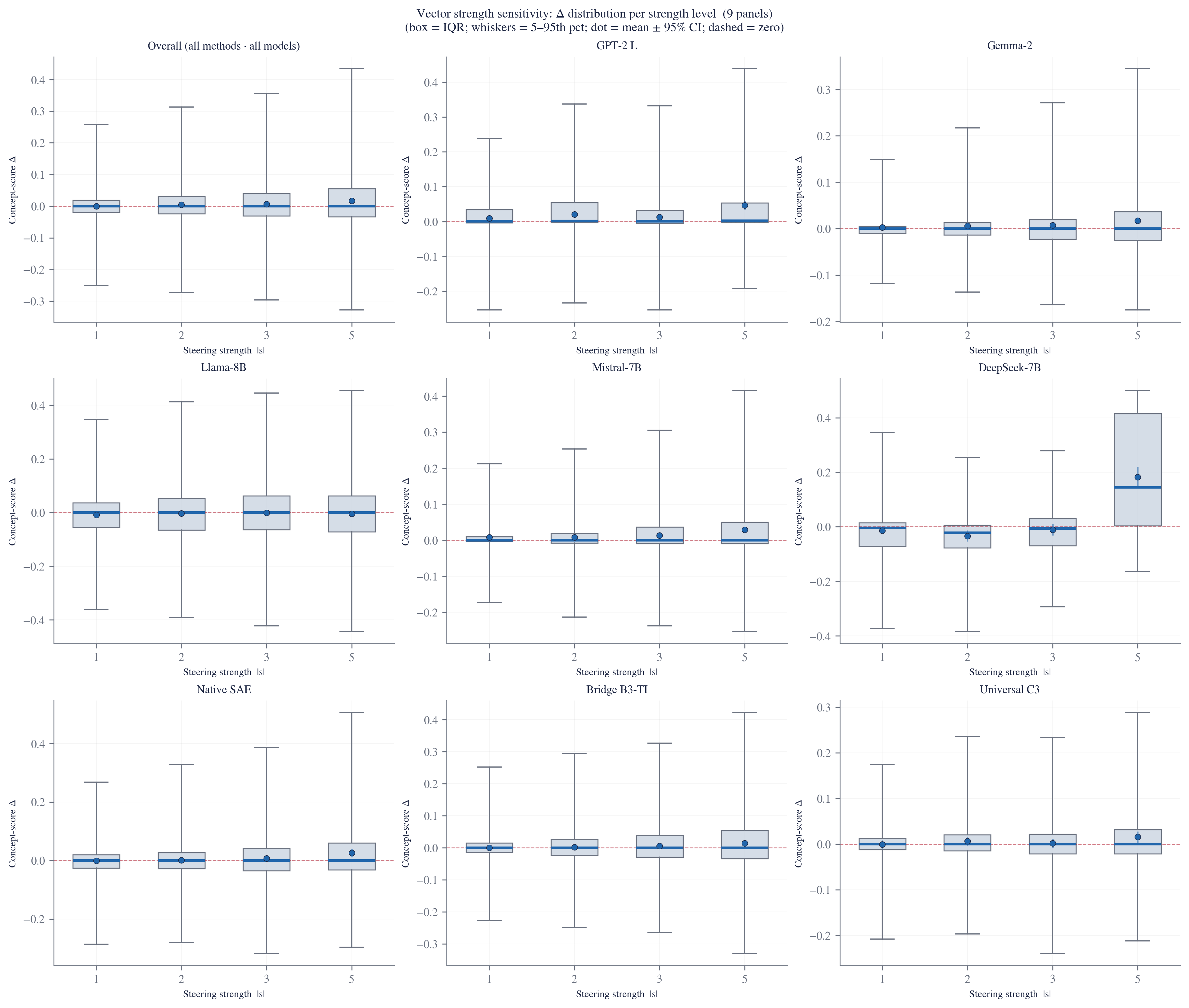}
\caption{Vector strength sensitivity: $\Delta$ distribution per strength level $|s| \in \{1,2,3,5\}$ (9 panels). Box = IQR, whiskers = 5th--95th percentile, dot = mean with 95\%~CI, dashed line = zero. DeepSeek-7B at $|s|=5$ shows markedly inflated variance.}
\label{fig:strength_boxplots}
\end{figure*}

\begin{figure*}[h!]
\centering
\includegraphics[width=\textwidth]{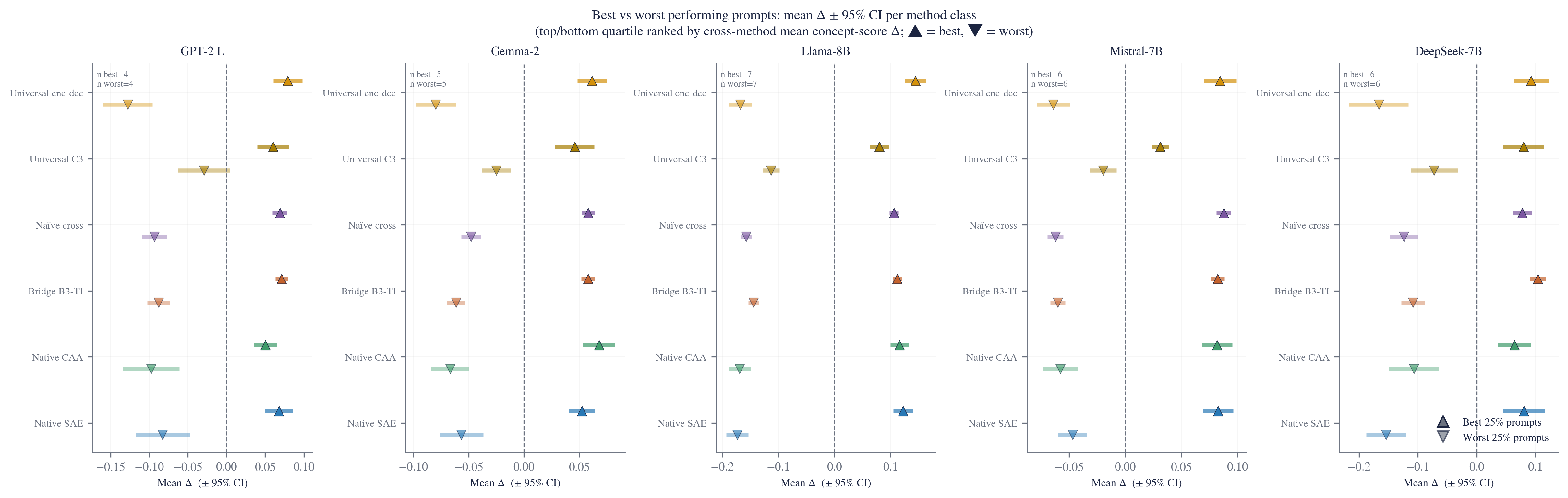}
\caption{Best vs worst performing prompts: mean~$\Delta \pm 95\%$~CI per method class (top/bottom quartile prompts ranked by cross-method mean~$\Delta$). $\blacktriangle$ = best quartile, $\blacktriangledown$ = worst quartile.}
\label{fig:best_worst}
\end{figure*}

\FloatBarrier
\section{Full Functional Results, Weak Domains}
\label{app:weak_domains}

Table~\ref{tab:supervised_full} (above, in Appendix~\ref{app:prompt_distrib})
provides mean delta, win rate, and $s=1$ delta for all methods and all five
models including the nine low-pass-rate domains.
The narrative below summarises the pattern for those nine domains.

The nine domains with B2 pass rate below 50\% --- \texttt{creative\_writing},
\texttt{legal}, \texttt{code\_python}, \texttt{code\_sql},
\texttt{math\_reasoning}, \texttt{code\_snippets}, \texttt{sentiment},
\texttt{science\_biomedical}, \texttt{question\_answering} --- show mixed or
negative Cohen's $d$ values in B2 (Table~\ref{tab:domains}), indicating that
the SAE features selected for those domains are not consistently discriminative
at the corpus level. For these domains, all methods (native SAE, B3-TI, Univ)
produce mixed signed deltas across models: no method is systematically above
50\% win rate when averaged over the full 30-prompt set.
This is mechanistically expected: concept discriminability in feature space
(Cohen's $d$) is a precondition for steerability. Where $d < 0.4$, the
steering vector direction is noisy and its injection does not reliably shift
output distribution toward the target register.
The distinction between the six strong domains (Table~\ref{tab:domains}, bold)
and the nine weak domains is therefore grounded in a measurable upstream
property of the feature representation, not a post-hoc selection on evaluation
outcome. All results are included in the released evaluation files
(\texttt{full\_eval\_results.jsonl}) for completeness and reproducibility.

\FloatBarrier
\section{Future Work}
\label{app:future}

The highest-priority deferred experiments are the preregistered
\textbf{disjoint-corpus ablation} (independent corpus subsets per model, to
rule out shared-statistics as the driver of alignment) and
\textbf{T3 activation-patching causal tracing} (inserting a steering vector
at one layer and patching the residual stream at subsequent layers to trace
the causal path, which would upgrade the T2 bidirectionality evidence to
mechanistic circuit-level attribution).

\textbf{Multi-layer injection.}
All experiments use single-layer injection at $\approx$50\% depth for clean
comparison with prior work. Sweeping injection depth and aggregating across
layers may reduce prompt-sensitivity and improve transfer to GPT-2-large.

\textbf{Representation-side questions.}
It remains open whether SAE architecture choices (TopK vs.\ ReLU, expansion
factor, $k$ schedule) meaningfully affect the cross-model alignment signal,
and whether the 15-concept supervised label set is a bottleneck for
feature-pair recall.

\textbf{Model family extensions.}
Mamba, mixture-of-experts, and encoder-decoder architectures are priority
extensions (the C3-EncDec cone collapse is a candidate mechanism to study in
cross-architecture crosscoders~\citep{lindsey2024crosscoders}); non-English
and multilingual models are also out of scope for the current study.

\textbf{Closed-source models.}
All experiments require mid-layer activation access; closed-source frontier
systems (GPT-4, Claude, Gemini) lie outside current scope. Developing
output-space or black-box analogues of the alignment pipeline---via
paired-output representation probing or distillation into an open proxy---is
a necessary step before these results can generalise to closed-source
deployment contexts.

\section{LLM-Judge Validation}
\label{app:llm_judge}

As a qualitative complement to DeBERTa concept-score deltas, a sample of
650 steered outputs was submitted to Claude (\texttt{claude-haiku-4-5},
temperature 0) with a forced-choice prompt: given the original unsteered
output and the steered output, the judge rates whether the steered output
more strongly expresses the target concept (win) or not.
Table~\ref{tab:llm_judge} reports Claude judge win rates on the 14 qualifying
prompts at two prompt-set sizes.

\begin{table}[h]
\centering
\caption{Claude judge win rate (\% steered output rated as expressing target
concept more strongly) on qualifying prompts ($\star$ = full hierarchy prompts only;
``Top-14'' = all qualifying prompts). Only cells with $\geq$3 valid comparisons
are shown; others are marked ---. Note: the LLM judge is a holistic quality
probe; it is not used to confirm statistical significance.}
\label{tab:llm_judge}
\small
\begin{tabular}{lrrrrrr}
\toprule
Method & deepseek & gemma & gpt2 & llama & mistral & Mean \\
\midrule
\multicolumn{7}{l}{\emph{Top-5 qualifying prompts}} \\
Native CAA  &  0.0 &  0.0 &  --- &  0.0 &  --- &  0.0 \\
B3-TI       &  --- & 25.0 &  0.0 &  --- & 50.0 & 25.0 \\
Naive       &  --- & 100.0&  --- & 50.0 &  --- & 75.0 \\
C3-Dec      & 100.0& 50.0 &  0.0 & 50.0 & 100.0& 60.0 \\
C3-EncDec   &  --- &  --- &  0.0 &  --- &  0.0 &  0.0 \\
\midrule
\multicolumn{7}{l}{\emph{Top-14 qualifying prompts}} \\
Nat. SAE-dec  &  0.0 &  --- & 100.0&  --- &  --- & 50.0 \\
Native CAA  &  0.0 &  0.0 &  --- & 50.0 & 100.0& 37.5 \\
B3-TI       &  0.0 & 33.3 & 40.0 &  --- & 40.0 & 28.3 \\
Naive       & 40.0 & 33.3 &  0.0 & 14.3 & 25.0 & 22.5 \\
C3-Dec      & 50.0 & 25.0 & 12.5 & 40.0 & 62.5 & 38.0 \\
C3-EncDec   &  0.0 &  --- &  0.0 &  --- &  0.0 &  0.0 \\
\bottomrule
\end{tabular}
\end{table}

\begin{figure}[h!]
\centering
\includegraphics[width=\linewidth]{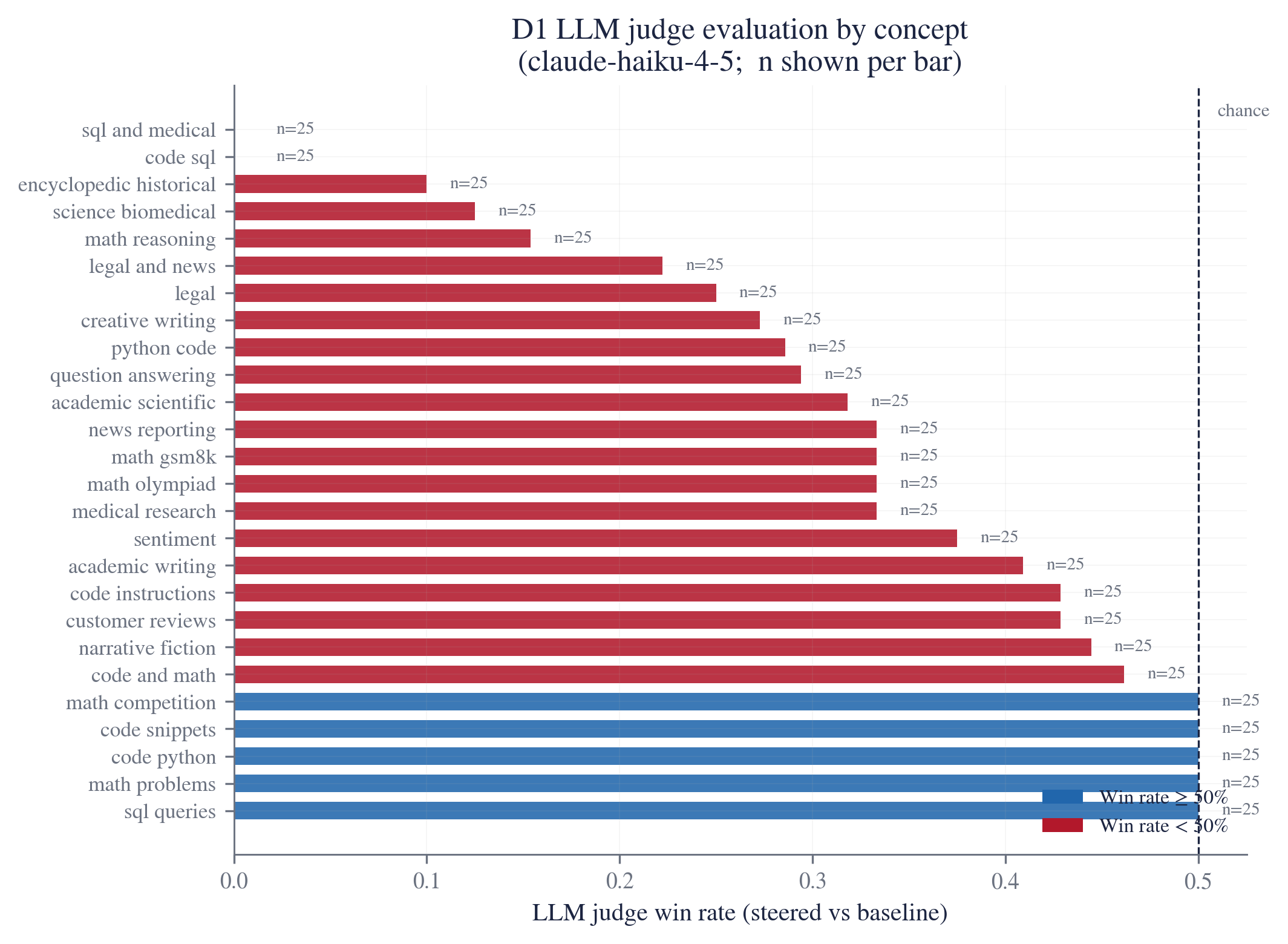}
\caption{LLM judge (\texttt{claude-haiku-4-5}) win rates: steered output vs baseline. Blue bars indicate win rate $\geq$50\%; red bars below chance. Sorted by win rate descending.}
\label{fig:judge_wins}
\end{figure}

The LLM judge and the DeBERTa concept-score delta are independent probes.
DeBERTa measures the shift in the token-level distribution over concept
labels; the LLM judge measures holistic output quality relative to the target
register. On the qualifying prompts, Univ~C3 achieves the highest LLM-judge
win rate (38.0--60.0\% depending on prompt subset), with deepseek showing
100\% on the top-5 prompts despite its low DeBERTa win rate, indicating that
when deepseek does produce on-topic output it is qualitatively strong.
The Pearson correlation between DeBERTa $\Delta$ and LLM-judge outcome is
$r = -0.004$ ($n = 228$), confirming the two metrics measure distinct aspects
of the steering effect and should not be treated as interchangeable.
All Claude judge prompts and raw outputs are stored in
\texttt{results/llm\_judge\_results\_prod.jsonl}.

\FloatBarrier
\section{Discarded Approaches}
\label{app:discarded}

\paragraph{Run 1: Hungarian pair extraction.}
An earlier run used \texttt{scipy.linalg.linear\_sum\_assignment} (Hungarian
method) for pair extraction, yielding 1,823 pairs with a B2 pass rate of 68.4\%.
This inflated result arose from three sources: (i) Hungarian forces global 1:1
coverage including weak pairs, but pre-selection of the top-$K$ features by SAE
label confidence guaranteed high discriminativity before B2 ran; (ii) the Cohen's
$d$ mean was inflated from 0.423 (Run 2) to 1.213 (Run 1) by this pre-selection;
(iii) gpt2-large cross-scale pairs appeared to pass at 65--79\% due to undirected
pooling that masked the scale gap.
All main-paper results use Run 2 (MNN extraction, 3,308 directed pairs, 45.5\%
overall pass rate).
The 45.5\% is not a regression; it is a more honest estimate that correctly
separates three scale-tier populations.
The correct comparison is Run 2 7B$\leftrightarrow$7B (48.9\%) vs Run 1
same-scale pairs (43--54\% after bias correction)---consistent.

\paragraph{HDBSCAN without UMAP (C2 Runs 1--2).}
Two early C2 runs applied HDBSCAN directly to the 512-d shared concept space.
Run 1 (min\_cluster\_size=20): 58\% noise.
Run 2 (min\_cluster\_size=50): 65\% noise (worse, due to over-merging).
Root cause: the curse of dimensionality causes pairwise distances in 512-d cosine
space to concentrate near their mean, eliminating density gradients required by
HDBSCAN~\citep{beyer1999nearest}.
Fix: UMAP reduction to 30d reduced noise to 15.8\% (Run 3, reported in main paper).

\paragraph{B3 SAE-Feature Locator and B3 Guided CAA (insufficient feature coverage).}
Two B3 strategies that operate directly on the B2-validated pair list
degenerated and are excluded from the main evaluation.

The \emph{B3 SAE-Feature Locator} constructs a target steering vector
by retrieving the target-model SAE feature that is B2-validated as
corresponding to each of the guide's top-3 concept features, then
accumulating confidence-weighted target decoder columns.
The method fails because B2 yields only 12--85 validated pairs per directed
model pair and none of these contain any of the top-3 concept features for
any of the 15 evaluated concepts. The locator therefore falls back to
writing the guide's own decoder weights as the output, producing
\texttt{sae\_decoder\_vector} entries numerically identical to the native A5
SAE-decoder vector for all 20 pairs $\times$ 15 concepts = 300 evaluation
cells. The root cause is architectural: B2-passing pairs are MNN-selected
for corpus-level cosine alignment, not for concept-label membership; the
probability that any specific concept's top-concept features fall in the
MNN-matched set is substantially lower than the 45.5\% aggregate pass rate
suggests.

The \emph{B3 Guided CAA} produces a mean-difference vector from the target
model's residual streams, split using the \emph{guide}'s concept activity (above
vs.\ below median on that concept's top features) rather than corpus labels.
A code defect prior to 2 May 2026 caused this path to duplicate the native
CAA vector; after correction \texttt{caa\_cross\_vector} is genuinely distinct,
but achieves no consistent advantage over native vectors in preliminary
evaluation at the tested strength range.
Both B3 BAL vectors are retained in the release under
\texttt{cross\_model\_steering\_vectors\_bal.json} as an audit trail but are not
included in any reported comparison.

\section{Compute Budget}
\label{app:compute}

\begin{table}[h]
\centering
\caption{Estimated A100-80GB GPU hours per experiment step.}
\label{tab:compute}
\small
\begin{tabular}{llcc}
\toprule
Step & Description & A100s & Wall time \\
\midrule
A2 & Activation extraction (all 5 models) & 1$\times$5 & $\sim$2h/model \\
A3 & SAE training (all 5 models)           & 1$\times$5 & $\sim$3h/model \\
A4 & Feature labelling                     & 1$\times$5 & $\sim$20min/model \\
B1 & Alignment MLP bridges (20 pairs)      & 1$\times$20 & $\sim$20min/pair \\
B2 & Validation (3,308 pairs)              & 1 & $\sim$60min \\
C1 & Global MLP training (200 epochs)      & 8 & 66 min \\
C2 & Universal concept discovery           & 1 & $\sim$15 min \\
C3 & Universal vector build                & 1 & $\sim$5 min \\
D1 & Functional evaluation (5 models)      & 5 (parallel) & $\sim$66 min \\
\midrule
\multicolumn{2}{l}{Total (reported experiments)} & \multicolumn{2}{c}{$\approx$52 A100-hours} \\
\bottomrule
\end{tabular}
\end{table}

\section{Released Assets}
\label{app:assets}

All large binary assets are released on Hugging Face:
\begin{center}
  \url{https://huggingface.co/datasets/ayushi-agarwal/universal-steering}
\end{center}
The complete pipeline source code is released on GitHub:
\begin{center}
  \url{https://github.com/ayushi-agarwal/universal-steering}
\end{center}

Table~\ref{tab:assets} summarises the released artefacts, their locations
within the dataset repository, and applicable licences.

\begin{table}[h]
\centering
\caption{Released artefacts. All are hosted at
\texttt{ayushi-agarwal/universal-steering} on Hugging Face unless marked
\textit{(git)}. Licence applies to the artefact itself; use is additionally
constrained by the upstream model licence where noted.}
\label{tab:assets}
\small
\begin{tabular}{llll}
\toprule
Artefact & HF folder & Size & Licence \\
\midrule
Domain-labelled corpus (394,508 passages) & \texttt{data/} & $\sim$2\,GB & CC-BY-SA-4.0 \\
Residual-stream activations (5 models) & \texttt{activations/} & $\sim$200\,GB & Per model \\
SAE checkpoints (5 models) & \texttt{saes/} & $\sim$1\text{--}5\,GB & Per model \\
4,100 feature label JSONs & \texttt{features/} \textit{(git)} & $\sim$10\,MB & Apache 2.0 \\
20 MLP alignment bridges (67M params each) & \texttt{alignment/} & $\sim$5\,GB & Per model \\
75 native steering vectors ($5\times 15$) & \texttt{steering/} & $\sim$80\,GB & Per model \\
300 cross-model vectors (B3-TI) & \texttt{steering/} & (included above) & Per model \\
Global concept MLP checkpoint & \texttt{universal/} & $\sim$1.3\,GB & Apache 2.0 \\
55 universal steering vectors ($11\times 5$) & \texttt{universal/} & (included above) & Apache 2.0 \\
B2 validation results (3,308 pairs) & \texttt{results/} & small & Apache 2.0 \\
Evaluation outputs (D1) & \texttt{results/} & small & Apache 2.0 \\
Pipeline source code & GitHub (ayushi-agarwal/universal-steering) & --- & Apache 2.0 \\
\bottomrule
\end{tabular}
\end{table}

Reproduction instructions are provided as \texttt{README.md} in the code
repository with exact commands and expected outputs for each pipeline step.
All randomness uses \texttt{seed=42} throughout.
Large binary assets are downloaded automatically by running
\texttt{python pipeline/a1\_download\_data.py} (requires a Hugging Face token
and acceptance of per-model terms).

\section{Asset Licences}
\label{app:licences}

Model licences: GPT-2 (MIT), Mistral-7B-v0.3 (Apache 2.0), LLaMA-3.1 (LLaMA 3
Community Licence, permits research and commercial use below 700M MAU),
Gemma-2-2B (Gemma Terms of Use, permits research), DeepSeek-LLM-7B (DeepSeek
Licence, permits research).
Dataset licences range from MIT (GSM8K, MetaMathQA) to CC-BY-SA-3.0 (Yelp,
Wikipedia); the derived corpus is released under CC-BY-SA-4.0.

\section{Use of Large Language Models}
\label{app:llm_usage}

Claude was used in three distinct roles in this pipeline, using different model
versions matched to each task's complexity and cost requirements.

\textbf{Feature labelling (A4) --- Claude \texttt{claude-sonnet-4-5}:}
The top-10 activating corpus passages per SAE feature were submitted to
\texttt{claude-sonnet-4-5} (temperature~0) with a prompt requesting a 2--4-word
domain label in \texttt{snake\_case} format; no few-shot examples were included
to avoid priming toward the 15 training domains. Sonnet was selected here
because label quality is directly upstream of B1/B2 alignment statistics.

\textbf{Universal concept naming (C2) --- Claude \texttt{claude-sonnet-4-5}:}
The top-5 activating passages per HDBSCAN cluster were submitted to
\texttt{claude-sonnet-4-5} (temperature~0) with an open-ended naming prompt;
out-of-corpus compound-domain examples (e.g., \texttt{sql\_and\_legal}) were
included to encourage cross-domain labels and prevent single-domain priming.

\textbf{LLM judge (D1 evaluation) --- Claude \texttt{claude-haiku-4-5}:}
650 steered outputs were evaluated by \texttt{claude-haiku-4-5} (temperature~0)
using a forced-choice prompt: given the original unsteered output and the steered
output, the judge rated which more strongly expresses the target concept.
Haiku was selected for the judge role because of the large evaluation volume
(650 comparisons) and because judge precision requirements are lower than
labelling precision --- the holistic quality comparison is binary and does not
require fine-grained semantic reasoning.

All prompts and raw Claude outputs are stored verbatim in the released feature
label JSON files (one file per model), the universal concept JSON file, and
\texttt{results/llm\_judge\_results\_prod.jsonl} respectively.
No LLM output was used in threshold selection, statistical inference, or
result interpretation; all such decisions were made by the authors.
Claude (\texttt{claude-sonnet-4-5}) was additionally used for grammar revision
of the manuscript.

\end{document}